%% file: acl_latex.tex
\documentclass[11pt]{article}

\PassOptionsToPackage{hyphens}{url}  % long URLs in the references break at hyphens
\usepackage[preprint]{acl}

\usepackage{times}
\usepackage{latexsym}
\usepackage{multirow}

\usepackage[T1]{fontenc}
\usepackage[utf8]{inputenc}
\usepackage{mathtools}
\usepackage{graphicx}

\usepackage{microtype}

\usepackage{inconsolata}

\usepackage{graphicx}
\usepackage{booktabs}
\usepackage{tabularx}
\usepackage{xcolor}

\title{MExAS: Measuring Exhaustive Agentic Search}

\author{
  Lyuxin David Zhang$^{1}$ \qquad
  Dan Roth$^{1,2}$ \qquad
  Tomer Wolfson$^{1}$ \\[4pt]
  $^{1}$University of Pennsylvania \quad
  $^{2}$Oracle AI \\[3pt]
  \texttt{\{davidzlx, danroth, wolfsont\}@engineering.upenn.edu} \\
}

\newcommand\dataset{\textsc{MExAS}}

\newif\ifcomments
\commentstrue
\ifcomments
    \newcommand\tomer[1]{\textcolor{magenta}{[TW: #1]}}
    \newcommand\david[1]{\textcolor{blue}{[DZ: #1]}}
    \newcommand\dan[1]{\textcolor{teal}{[RT: #1]}}
\else
    \newcommand\tomer[1]{}
    \newcommand\david[1]{}
    \newcommand\dan[1]{}
\fi

\begin{document}
\maketitle

\input{content/abstract}

\input{content/introduction}

\input{content/benchmark}

\input{content/evaluation}

\input{content/analysis}

\input{content/related_work}

\input{content/conclusion}

% \section*{Acknowledgments}

% Bibliography entries for the entire Anthology, followed by custom entries
%\bibliography{custom,anthology-overleaf-1,anthology-overleaf-2}

% Custom bibliography entries only
\bibliography{custom,anthology}

\input{content/appendix/main}

\end{document}

%% file: content/abstract.tex
\begin{abstract}
Language model-powered agents have become a popular approach to querying document corpora, providing users with easy access to vast amounts of data.
However, prior work on agentic search has largely overlooked the problem of \emph{exhaustive search}: identifying every relevant answer within a large-scale corpus. This setting is particularly important in high-stakes domains such as law and medicine, where failing to retrieve even a single relevant legal case or treatment option can have serious consequences.
To address this gap, we introduce \dataset{}, a benchmark for systematically evaluating the exhaustive search capabilities of agents. \dataset{} introduces exhaustive search tasks over 10 corpora in fact-checking, programming, law, and medicine. Using \dataset{}, we evaluate the performance of frontier search agents built with the Codex and Claude Code harnesses. Our results show that all agents struggle on exhaustive search tasks, with the top performing Opus~5.5 reaching just 64.4\% F1. Importantly, our analysis reveals misleading behavior, as Codex agents rarely acknowledge that their results may be incomplete even when reviewing less than 50\% of the corpus.
These findings underscore the limitations of existing agents in tackling exhaustive search tasks.
Our codebase and data are publicly available.\footnote{\url{https://huggingface.co/datasets/CogComp/MExAS}}
\end{abstract}

%% file: content/introduction.tex
\section{Introduction}

\begin{figure}[t]
  \centering
  \includegraphics[width=0.82\columnwidth]{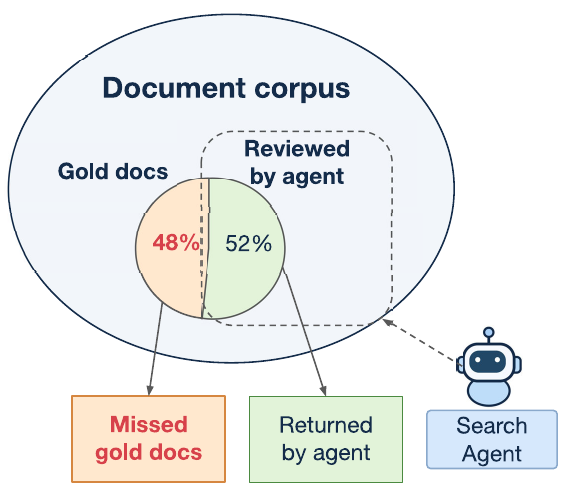}
  \caption{An agent that struggles to search exhaustive, failing to find 48\% of all relevant documents.}
  \label{fig:motivation}
  \vspace{-0.43cm}
\end{figure}

Recent advances in Large Language Models (LLMs) have enabled the development of autonomous agents capable of writing and executing code, searching the web, and synthesizing information from multiple sources  \cite{yang2024sweagent,zhou2024webarena,wei2025browsecomp, karten2026primeagentselfimprovingrlm}.

An important application of LLM-powered agents is their ability to query large document corpora accurately and efficiently, allowing users to explore vast collections of information simply by asking questions. Recent research on agentic search has increasingly shifted its focus from answering complex questions whose evidence is concentrated in a handful of documents \cite{chen2025browsecomp, wong2026widesearch} to tackling tasks that require synthesizing information across dozens or even hundreds of documents \cite{wolfson2026monaco,paul2026deepsynth,pmlr-v306-zeng26g, jang2026odysseys}.  

However, what if a user's search task requires examining every document in a very large corpus? Consider, for example, a physician reviewing the medical literature to identify drugs associated with a particular adverse reaction, or a paralegal searching for all court opinions that reference a given set of cases. In such high-stakes domains, overlooking even a small number of relevant documents can have significant consequences.
Therefore, when delegating these tasks to automated agents, it is essential to understand the extent to which their search results are truly \emph{exhaustive}.

An exhaustive search task requires the search agent to retrieve \emph{every} document in the corpus that satisfies a specified condition. Such conditions may be expressed as syntactic patterns, for example, finding, ``Which individuals mentioned went to college?'' can be solved using the pattern: [PERSON] \emph{was educated at} [LOCATION] \emph{college}. These patterns can often be inferred from a small sample of the corpus \cite{hearst-1992-automatic, goldberg2023two}.
However, exhaustive search tasks may also involve identifying implicit properties of target documents, such as ``identifying all contradictory claims in the corpus'' \cite{tchuindjo2026obliq, singhal2026no}. An ideal search agent should be capable of addressing both types of challenges without relying on manually constructed pipelines or direct user steering. It should produce results that are both complete and cost-optimal, adapting its search strategy to the structure of the corpus.

To study these capabilities, we introduce \dataset{}, a benchmark for \textbf{M}easuring \textbf{Ex}haustive \textbf{A}gentic \textbf{S}earch. \dataset{} adapts 10 datasets of large document corpora (ranging from 4.6K to 291K documents) across four domains—fact-checking, programming, law, and medicine—into exhaustive search tasks in which the target set of documents is sparse ($\sim$5\% of the corpus). The benchmark contains 100 query-corpus pairs, some of which can be solved using lexical patterns, while others require the LLM to directly reason over documents in order to determine their relevance.

We use \dataset{} to evaluate three frontier search agents: GPT-6.1-Sol and GPT-6-Astra using Codex as their agentic harness, and Claude Opus~5.5, using Claude Code. Our results reveal a 30-point gap in F1 scores between document classification (i.e., identifying relevant documents) and retrieving those documents from the corpus---highlighting the challenge that  exhaustive retrieval poses. All agents fail to perfectly retrieve all target documents, with the top performing Opus~5.5 reaching 64.4\% F1.

Codex-based agents demonstrate a particularly low recall (62.8--69.6\%). More important, our analysis reveals these agents rarely acknowledge in their outputs that their results may be incomplete, even while reviewing less than 50\% of the corpus. Figure~\ref{fig:motivation} illustrates this behavior, where an agents reviews only a subset of the corpus results in many of the target documents being missed.

While Opus~5.5 present higher recall at 81.9\% it comes at a much larger cost---consuming 2.56$\times$ as many tokens as those in the corpus itself compared with roughly 0.06--0.08$\times$ for Codex. Furthermore, we evaluate the cost-effectiveness of frontier agents by comparing them against a baseline agent hard-coded to exhaustively classify the entire corpus. Despite relying on a substantially weaker LLM (GPT-6-Luna), our baseline achieves recall comparable to Opus~5.5 (80.8\%) while using 1.6$\times$ fewer tokens. This proof of concept demonstrates that frontier search agents remain far from cost-optimal.

Lastly, we analyze agent trajectories and reveal that Codex and Claude based agents diverge in terms of their search strategies. Codex-based agents rely on rule-based patterns resulting in low recall and results that are far from exhaustive. Claude distributes the corpus review across multiple sub-agents, achieving higher recall but at a significantly increased cost.
Ultimately, both approaches fail to present a cost-optimal solution for \dataset{}, highlighting the need to build better agents for exhaustive search.

Our paper contributions are as follows:

\begin{itemize}
    \item We build \dataset{}, an evaluation benchmark designed to assess the accuracy and cost of agents on exhaustive search tasks (\S\ref{sec:benchmark}).
    
    \item We demonstrate that frontier agents struggle with exhaustive search over large document corpora, exhibiting limitations in both recall and precision, as well as challenges in optimizing costs across different corpora (\S\ref{sec:evaluation}).

    \item We analyze agent trajectories and identify two distinct search strategies, each associated with a different harness (Codex or Claude), highlighting the trade-offs between them in terms of search performance and cost (\S\ref{sec:analysis}).
\end{itemize}

Finally, we release all corpora, queries, and agents trajectories, with the hope that \dataset{} will aid research on new agentic search approaches capable of more effectively tackling exhaustive search over tens of thousands of documents.

%% file: content/benchmark.tex
\section{The \dataset{} Benchmark}
\label{sec:benchmark}

\subsection{Task Definition}
\label{sec:task}

\begin{figure*}[t]
    \centering
    \includegraphics[width=0.93\textwidth]{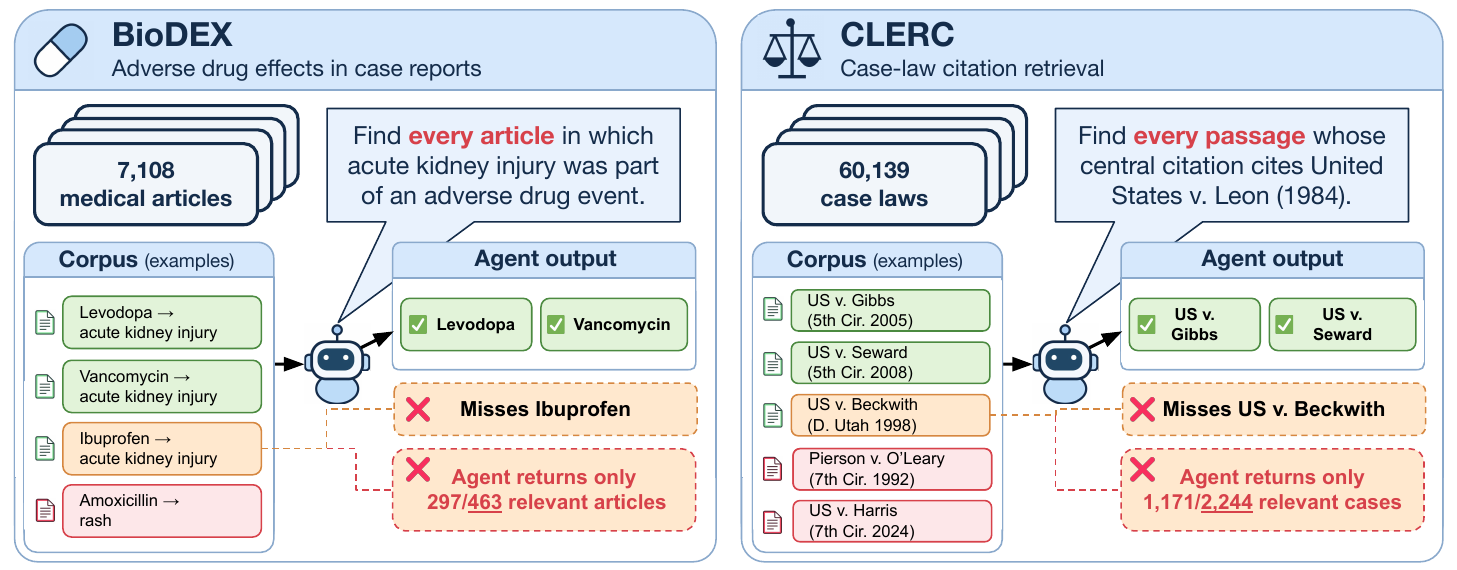}
    \caption{Two exhaustive search tasks from \dataset{}. The BioDEX task asks for medical papers that report a particular drug reaction (kidney injury). In CLERC the agent must return all passages whose masked citation refer to a particular legal case (US v. Leon).
    Also presented are the results of Codex with
    GPT-6.1-Sol on these tasks.}
    \label{fig:two_mexas_examples}
\end{figure*}

Given a corpus of documents $D=\{d_1,\ldots,d_N\}$ and a natural-language query $Q_D$, exhaustive search requires finding the complete answer set $D_A$ supported by the corpus.
Given an agent's answer set $\hat{D}_A$ to query $Q_D$, we use $D_A$ to compute the agents precision, recall and F1 score.
% \[
% P=\frac{|\hat{A}\cap A|}{|\hat{A}|},
% R=\frac{|\hat{A}\cap A|}{|A|},
% F_1=\frac{2PR}{P+R}.
% \]

\subsection{Benchmark Construction}
\label{sec:construction}
\dataset{} is comprised of a diverse set of real-world domains spanning fact-checking, programming, law and medical corpora. For each domain, we use datasets of human written document corpora, introduced by prior work, and modify these datasets to fit an exhaustive search setting. Overall, we introduce 10 search tasks which are grouped into 4 domains: fact-checking, programming, law and medicine.

For each task, the search agent is prompted to retrieve a subset of the document corpus which answer its search query. To further expand \dataset{}, we use bootstrap sampling \cite{efron1979bootstrap} and create 10 examples for each search task $Q_D^1, ..., Q_D^{10}$ by sampling $D_1, ..., D_{10}$, which are random subsets of $D$ containing $\frac{|D|}{2}$ documents.
We deliberately keep the set of answer documents to be relatively small, between 3.7--8.4\% of $|D|$ for each dataset. It is meant to reflect real-world search tasks where the relevant information may be highly dispersed \cite{goldman-etal-2024-really}. Table~\ref{tab:benchmarks} lists the 10 datasets used in \dataset{} and their statistics. 

\input{content/tables/tab_benchmarks.tex}

\subsection{Evaluation Tasks}
\label{sec:eval_tasks}
Figure~\ref{fig:two_mexas_examples} presents two examples one from the BioDEX and one from the CLERC corpora. Irrelevant documents are in red, target documents returned by the agent are in green while the target documents that were missed by the agent are in orange. In both examples the agent misses 36\% and 48\% of all relevant documents.

Next, we briefly describe each of our 10 evaluation tasks, while task-specific prompts and data filtering details are presented in Appendix~\ref{app:tasks}--\ref{app:prompts}.

\subsubsection{Fact Checking}
\label{subsec:eval_tasks_fact_checking}
\paragraph{NLI.} We frame natural language inference (NLI) as an exhaustive search task using data from the SNLI~\citep{bowman-etal-2015-large}, MultiNLI~\citep{williams-etal-2018-broad}, ANLI~\citep{nie2020adversarial} and VitaminC~\citep{schuster2021get} datasets. These datasets contain pairs of statement and passage where the statement either contradicts the text in the passage, or not. For each passage we select a random statement that either contradicts or supports it, so that there is only a single relevant statement per passage. Each statement-passage pair is considered a single document resulting in a corpus of 291K examples. The exhaustive search task is finding all pairs of contradicting statements in the corpus (Table~\ref{tab:q1}), of which there are 11.6K gold documents. 

\paragraph{HoVer.} A multi-hop fact verification dataset \cite{jiang2020hover}. HoVer contains factual 26K claims, where each claim is paired 8 Wikipedia paragraphs with 2--4 of the paragraphs serving as evidence that either supports or does not support the claim. Since HoVer claims typically have multiple variants, we filter the original dataset, keeping only a single variant per claim. We treat each claim and its paired Wikipedia paragraphs as a unique document, resulting in 7,248 documents. The task is to find all claims that are not supported by their relevant paragraphs (Table~\ref{tab:q1}), which we set at 290.

\paragraph{FEVEROUS.} A fact verification dataset over Wikipedia tables and text \cite{aly2021feverous}. Each claim in the data is paired with evidence sentences and/or table cells. The evidence either supports the claim, refutes it or does not provide enough information to reach a verdict. The search task asks the agent to return all of the refuted claims (Table~\ref{tab:q1}), which we down sample to be for 4\% of the documents, resulting in a corpus of 50K documents.

\subsubsection{Programming}
\label{subsec:eval_tasks_programming}
\paragraph{Codeforces + CodeNet.}
We combine two popular programming problems into an exhaustive search corpus. Codeforces~\citep{penedo2025codeforces} contains 10K programming problems paired with 12.6M submitted solutions, while CodeNet~\citep{puri2021codenet} has 4,053 problems with 13.9M submissions. Each submission is labeled as either \emph{accepted} or as number of different error types. We use only submissions with the \emph{accepted}, \emph{wrong} and \emph{runtime error} labels. Overall, we keep 6,109 pairs of problems + solutions from both datasets, which serves as our document corpus. The task is to return all programs whose solution was not accepted (Table~\ref{tab:q1b}), which we set to be 5\% of the corpus.

\subsubsection{Law}

\paragraph{CaseHOLD.} 
A multi-choice QA dataset in the legal domain. CaseHOLD \cite{zheng2021does} contains 53K judicial decisions where each decision contains a legal citation to a previous relevant case together with a ``holding statement'' that explains the cited legal precedent. In CaseHOLD, the holding statement is masked and the task is to choose the correct one from 5 candidate holdings. Each document is a judicial decision excerpt paired with a single candidate holding, where we set 4.2\% of decisions being paired with an incorrect holding statement. 
The search task in \dataset{} is to return all court decisions with an erroneous holding (Tables~\ref{tab:q1b2}).
For the corpus, we filter it down to 50.5K documents, so that each holding appears as the correct holding in exactly one court decision.

\paragraph{CLERC.}
A legal case retrieval dataset by \citet{hou2025clerc} with 1.84M cases chunked into multiple passages. The middle of each passage has a ``central citation'', referencing to a previous legal case that the current case relies on. In \dataset{} we mask the central citation, as well as its abbreviation and case ID mentions within the passage. Due to budget constraints, we maintain all the passages whose central citation belongs to one of 77 unique cases, all concerning a similar topic. This results in a corpus of 60K passages, which includes all passages where the the original 77 cases are the central citation, as well as the full text of 4 target cases mentioned in the prompt. The exhaustive search task is to return all passages whose masked citation refers to one of 4 unique cases mentioned in the task prompt (Table~\ref{tab:q1c1}). There are 2,244--5,063 answer passages for each question, which are 3.7--8.4\% of the corpus.

\paragraph{CUAD.} A legal contract understanding dataset by \citet{hendrycks2021cuad}. CUAD contains data from 510 contracts where legal experts annotated each contract with 41 clause categories. The resulting dataset contains over 13K annotated clauses. We convert CUAD into a search task by treating each contract clause as a distinct document, while asking the agent to return all clauses which belong to a particular category, described in its prompt (Table~\ref{tab:q1c1}). Following de-duplication of clauses and removal of overly short ones (fewer than 8 words), we are left with a corpus of 8.3K clauses.
The agent is tasked with returning all clauses belonging to 4 categories out of of the 41, each category having around 400 clauses (5\% of the corpus).

\subsubsection{Medicine}

\paragraph{Evidence Inference 2.0.} 
A dataset for labeling clinical trial report outcomes \cite{deyoung2020evidence}. It contains reports of Randomized Control Trials (RCTs) from 3,346 PubMed articles. Each report describes a medical intervention given to a trial group while a ``comparator'' (e.g. placebo) was given to another group. The trial also reports the \emph{outcome} of the treatment within the two groups. Doctors then labeled for each intervention whether it led to a \emph{significant increase}, \emph{significant decrease}, or
produced \emph{no significant difference} in the outcome,
relative to the comparator---resulting in 12.5K annotations. In \dataset{}, each document is a single comparison paired with the full text of its corresponding trial report which, following data filtering, results in 8.16K documents (Appendix~\ref{app:task-evinf}). The agent is tasked with returning all the comparisons whose outcome is \emph{decreased} (Table~\ref{tab:q2}), which are set at 4.7\% of the documents.

\paragraph{MedQA+.} This is a collection of multi-choice medical QA datasets. We combine examples from the MEDEC \cite{abacha2025medec} dataset for medical error detection together with three other MedQA datasets \cite{hendrycks2020measuring,jin2021disease,pal2022medmcqa}. We convert these datasets to a medical error detection setting, by changing the question and one of the candidate answers into a statement that is appended to the end of the patient's description. If the candidate answer was incorrect, then the resulting text contains an error. The agent is then tasked with returning all clinical texts that contain an error (Table~\ref{tab:q2c1}). Following data filtering, the corpus contains 4.7K texts with 189 target documents (4.0\%).

\paragraph{BioDEX.}
A dataset for biomedical paper classification by \citet{d2023biodex}. It consists 19K biomedical papers that contain reports of patients that include their details, a set of
drugs taken by the patient, their dosage, and the adverse drug reactions experienced by the patient. In \dataset{} we frame this as a search task where the agent must find all the papers where the adverse drug reactions belong to one of 5 particular reactions (Table~\ref{tab:q2c1}). The corpus consists of 7.1K papers where the papers featuring the target adverse reactions constitute 4.1--6.5\% of the papers for each reaction.

%% file: content/tables/tab_benchmarks.tex
\begin{table*}[t]
\centering
\scriptsize
\setlength{\tabcolsep}{3pt}
\renewcommand{\arraystretch}{1.13}
\begin{tabularx}{\textwidth}{@{}
  >{\raggedright\arraybackslash}p{2.1cm}
  >{\raggedright\arraybackslash}p{4.4cm}
  c r r
  >{\raggedright\arraybackslash}X
  @{}}
\toprule
\textbf{Domain} & \textbf{Benchmark} &
\textbf{Documents} & \textbf{Tokens} &
\textbf{Answer \%} &
\textbf{Target Documents} \\
\midrule
Fact-checking & \textbf{NLI}~\textcolor{blue}{\citep{bowman-etal-2015-large,williams-etal-2018-broad,nie2020adversarial,schuster2021get}} & 290,956 & 15.5M & 4.0\% & Statements contradicted by their passage. \\
 & \textbf{HoVer}~\textcolor{blue}{\citep{jiang2020hover}} & 7,248 & 7.0M & 4.0\% & Multi-hop claims whose 2--4 evidence paragraphs contradict. \\
 & \textbf{FEVEROUS}~\textcolor{blue}{\citep{aly2021feverous}} & 50,505 & 7.2M & 4.0\% & Factual claims refuted by their relevant table or text.\\\midrule
Programming & \textbf{Codeforces + CodeNet}~\textcolor{blue}{\citep{penedo2025codeforces,puri2021codenet}} & 6,109 & 5.1M & 5.0\% & Code submissions rejected for either returning a wrong answer or a runtime error. \\\midrule
Law & \textbf{CaseHOLD}~\textcolor{blue}{\citep{zheng2021does}} & 50,579 & 12.7M & 4.2\% & Judicial decisions whose ``holding statement'' is incorrect, given the cited case. \\
 & \textbf{CLERC}~\textcolor{blue}{\citep{hou2025clerc}} & 60,139 & 25.3M & 3.7--8.4\% & All legal passages whose masked citation refers to one of 4 specific legal cases in the corpus. \\
 & \textbf{CUAD}~\textcolor{blue}{\citep{hendrycks2021cuad}} & 8,306 & 0.7M & 4.7--5.3\% & Contract clauses which belong to a particular category. \\\midrule
Medicine & \textbf{Evidence Inference 2.0}~\textcolor{blue}{\citep{deyoung2020evidence}} & 8,162 & 52.3M & 4.7\% & Trial reports where the medical intervention led to a ``decreased'' outcome relative to a comparison group. \\
 & \textbf{MedQA+}~\textcolor{blue}{\citep{hendrycks2020measuring,jin2021disease,pal2022medmcqa,abacha2025medec}} & 4,676 & 0.6M & 4.0\% & Clinical statements that contain a medical error. \\
 & \textbf{BioDEX}~\textcolor{blue}{\citep{d2023biodex}} & 7,108 & 25.7M & 4.1--6.5\% & Biomedical papers that report one of 5 pre-specified adverse drug reactions. \\
\bottomrule
\end{tabularx}
\caption{\dataset{} data statistics. For each of our 10 exhaustive search tasks we list their corpus:  number of documents, total length in tokens. We also list the share of target documents (Answer \%), shown as a range for CLERC, CUAD, and BioDEX which contain several target classes.
The full dataset construction is in Appendix~\ref{app:tasks}.}
\label{tab:benchmarks}
\end{table*}

%% file: content/evaluation.tex
\section{Experiments}
\label{sec:evaluation}

\begin{figure*}[t]
  \centering
  \includegraphics[width=\textwidth]{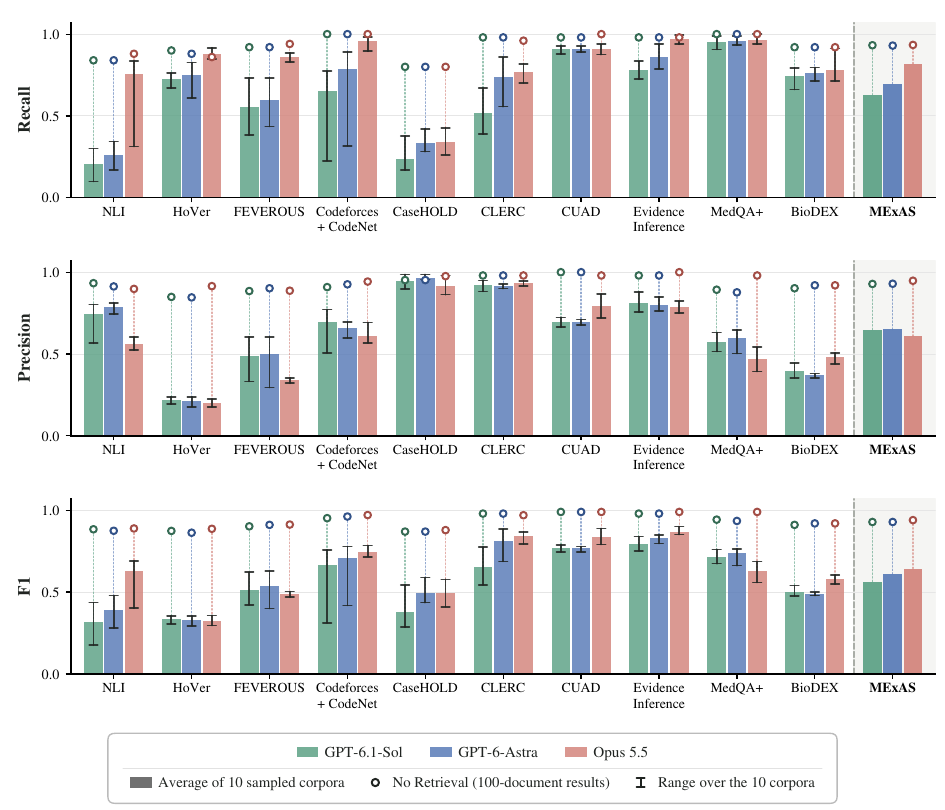}
  \caption{Agents performance gap on \dataset{} exhaustive search vs. their \emph{no-retrieval} scores. Filled bars represent the averaged agent performance over 10 corpus samples per task, while the open markers represent each agents' performance on the no-retrieval sample. The intervals indicate the highest and lowest scores in each of those runs. MExAS results (rightmost column) are the average scores over all 10 tasks. Detailed results listed in Table~\ref{tab:results}.}
  \label{fig:main_results}
\end{figure*}

Having introduced \dataset{}, we use it to evaluate three state-of-the-art search agents. We first describe our evaluation setup (\S\ref{sec:experiments:eval_setup}). In \S\ref{sec:experiments:results}, we show that agents struggle on exhaustive search tasks over large document corpora, achieving significantly lower performance than in an oracle retrieval setting. In \S\ref{sec:experiments:corpus_size}, we investigate the impact of corpus size on agent performance. Finally, in \S\ref{sec:experiments:cost}, we examine the cost-effectiveness of the different search agents and compare them against a baseline agent that exhaustively reads the entire corpus using a significantly weaker LLM.

\begin{figure*}[t]
  \centering
  \includegraphics[width=\textwidth]{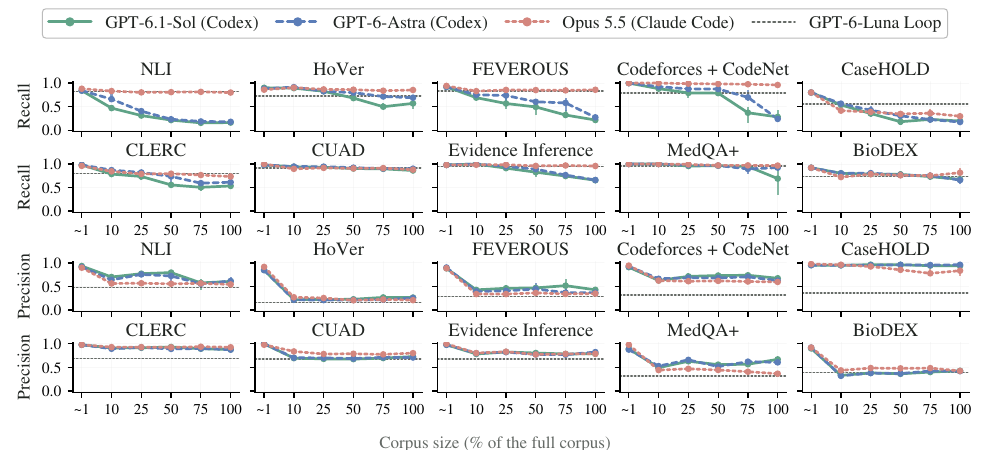}
  \caption{Agents precision and recall as we increase the corpus size. Full results are listed in Table~\ref{tab:sizes}. The leftmost point ($\sim$1\%) shows a single run on the 100-document results, while the 10\% to 100\% points present the averages score of 3 runs. The dotted line presents the results of the exhaustive GPT-6-Luna agent from \S\ref{sec:experiments:exhaustive_loop} (see Table~\ref{tab:luna}).}
  \label{fig:sizes-main}
\end{figure*}

\subsection{Evaluation Setup}
\label{sec:experiments:eval_setup}
We experiment with GPT-6.1-Sol~\citep{openai2026gpt61sol} and GPT-6-Astra~\citep{openai2026gpt6astra} using OpenAI's Codex~\citep{openai2025codex} agentic harness, and with Claude Opus 5.5~\citep{anthropic2026opus55} using Claude Code.\footnote{Claude sub-agents are restricted to Haiku~5.5 due to budget constraints.}
All LLMs are set to an ``extra-high'' reasoning level and granted shell access to search files, write code, and execute it. To avoid agents' potentially cheating, each run is executed in an isolated sandbox that blocks internet access and hides other repositories on the disk. The workspace contains only the corpus and a prompt asking the agent to write its formatted answers to \texttt{answer.txt}. The full task prompts are provided in Appendix~\ref{app:prompts}.

\subsection{Main Results}\label{sec:experiments:results}
Figure~\ref{fig:main_results} presents agents' performance on \dataset{}, with full results in Table~\ref{tab:results}. Following the description in \S\ref{sec:construction} each agent was evaluated on 100 runs in total: for each of the 10 search tasks, we randomly sampled 10 distinct document corpora, each comprising half of the total document set associated with that task. The results show that all three agents struggle on \dataset{}, with the top performing Opus~5.5 reaching an average F1 score of 64.4\% over all 100 examples. GPT-6.1-Sol demonstrates the lowest recall with an average of 62.8\%, far from perfectly returning all relevant documents.

A key question is whether the agents’ poor performance on \dataset{} stems primarily from the complexity of its search problem or whether the document classification task itself poses the main challenge. To disentangle the difficulty of document classification from that of the search problem in \dataset{}, we adopt a methodology similar to that used to study the \emph{retrieval--verification asymmetry} \cite{tchuindjo2026obliq}. Namely, we evaluated the agents on a random sample of 100 documents per task (1,000 documents in total), consisting of an equal number of gold documents (i.e., documents belonging to the target class) and irrelevant documents. Agents results on these \emph{no-retrieval} samples are represented by the open markers in Figure~\ref{fig:main_results}. All agents achieve very high performance, with average F1 scores ranging from 92.9--94.0\%, showing a large performance gap compared to the ``corpus search'' setting. Therefore, the drop in both recall and precision observed in the search setting suggest that exhaustively retrieving all relevant documents is considerably more challenging than correctly classifying individual documents.

\subsection{Varying the Corpus Size}\label{sec:experiments:corpus_size}
Having observed that searching for target documents in a large corpus harms agents performance, we examine whether their results further degrades as the corpus size grows. 

Figure~\ref{fig:sizes-main} presents the agents’ performance across six different \% of the full document corpus: $\sim$1\% (100 documents), 10\%, 25\%, 50\%, 75\% and 100\%. We run the agent three times on each of the five samples and present their averaged precision and recall. The full results are presented in Table~\ref{tab:sizes}. 

% recall
\paragraph{Recall.} Looking at the recall scores, the Codex-based agents display a drop in performance as the corpus size increase on all of the 10 search tasks. Their average recall drops significantly: from 80\% and 84.3\% over the 10\% subset to 48.4\% and 53.1\% when evaluated on the full corpus. In contrast, Claude Opus~5.5 scores much higher on recall, exhibiting a very modest performance drop as the corpus size grows (from 84.2\% to 81.2\%). This behavior can largely be attributed to the search strategy of Opus~5.5, which we further analyze in \S\ref{sec:analysis}. Nevertheless, all 3 agents' fail to achieve perfect recall, while the Codex-based agents are far from exhaustive, with their recall directly tied to the number of documents they are asked to review. 

Breaking performance by individual tasks, we observe that all agents achieve very high recall on CUAD and MedQA+. This is unsurprising, given that their document corpora span 0.6--0.7M tokens, and therefore fits around the same number of tokens as the underlying LLM's context window.

% precision
\paragraph{Precision.} 
Interestingly, all three agents perform roughly on par when it comes to precision. Precision scores do not significantly change as corpus size increases from 10\% to 100\% (62.1\% to 58.4\% for Opus). This is to be expected, since precision depends on how accurate the agent is in classifying the documents as the target documents. Increasing the total number of documents should not affect the per document classification performance, keeping the precision unchanged as well. We do note a decrease in precision compared to the 100 document sample ($\sim$1\% in Figure~\ref{fig:sizes-main}). This can largely be explained by the much lower prevalence of target documents compared to irrelevant ones (Tables~\ref{tab:benchmarks}), potentially resulting in more false positives, compared to the balanced 100 documents sample. To further assess the effect of false positives we present a detailed error analysis in Appendix~\ref{app:experiments:error_analysis}.

\paragraph{Exhaustive Search Agent.}\label{sec:experiments:exhaustive_loop}

We compare the performance of frontier agents against a brute-force baseline that exhaustively reads every document in the corpus. We use the cheaper GPT-6-Luna~\citep{openai2026gpt6luna}, with high reasoning, as the reader and compute its performance on each \dataset{} task over a 50\% sample of the full corpus. The full details are provided in Appendix~\ref{app:luna}.

The results of the exhaustive agent baseline are presented by the dotted line in Figure~\ref{fig:sizes-main}. Although the agent is evaluated only on a 50\% sample of the corpus, its exhaustive design leads us to expect consistent performance across different corpus sizes. In terms of recall, the exhaustive baseline outperforms or matches at least one of the frontier agents on 8 of the 10 tasks. Table~\ref{tab:cost-summary} compares the average performance on \dataset{} of the exhaustive baseline and the three frontier agents. The baseline achieves recall comparable to Opus~5.5 (80.8\% vs. 81.9\%), while substantially outperforming the recall of the Codex-based agents (62.8\% and 69.6\%). As expected, however, the exhaustive agent achieves lower precision, reflecting its use of a weaker underlying LLM (GPT-6-Luna) to reason over each document. 

In terms of runtime cost, we note that the exhaustive baseline consumes only 62.9\% of the amount of tokens consumed by Opus~5.5 (1.61$\times$ vs. 2.56$\times$, while having comparable recall. We discuss this cost-effective tradeoff next in \S\ref{sec:experiments:cost}.

\subsection{Cost-optimal Search}\label{sec:experiments:cost}

Table~\ref{tab:cost-summary} compares the agents in terms of their cost, LLM calls, tokens consumption and overall performance. The results are averaged over 10 distinct 50\% samples of \dataset{} resulting in 100 runs (i.e., the average results of 10 runs for each task). Exact cost breakdown is provided in Tables~\ref{tab:luna}--\ref{tab:tokens}, with LLM pricing described in \S\ref{app:cost}.

To assess how exhaustively an agent reviews the corpus, we report the ratio of the agent's total input, reasoning, and output tokens to the total corpus size, denoted as \emph{New input + output / corpus}. A ratio below 1 indicates that the agent has reviewed only a subset of the corpus. Ideally, an agent should achieve perfect recall while processing as few documents as possible.

We observe that Codex-based agents consume substantially fewer tokens than the corpus size, with token ratios ranging from 0.06--0.08$\times$. However, this appears to come at the expense of recall, with scores ranging from 62.8\% to 69.6\%. Claude Opus~5.5 achieves the highest recall (81.9\%), but consumes approximately 1.6 times as many tokens as the exhaustive search baseline while improving recall by only 1.1 points (81.9\% vs. 80.8\%). While the exhaustive baseline exhibits lower precision, its performance can be improved by applying a stronger agent to its outputs at a substantially lower cost than using Opus~5.5 on the entire corpus. As a proof of concept, our baseline demonstrates that, for exhaustive search tasks, even a simple strategy can achieve comparable or higher recall than frontier agents at a much lower cost (\$29 vs. \$317).

Overall, all agents exhibit a trade-off between cost and recall: they either incur substantial overhead or fail to achieve high recall. These findings highlight the need for more cost-efficient agents capable of performing exhaustive search effectively.

\begin{table}[t]
    \centering
    \footnotesize
    \setlength{\tabcolsep}{2.5pt}{\input{content/tables/tab_cost_summary.tex}}
    \caption{Comparing the exhaustive agent baseline to frontier agents in terms of cost, precision and recall. Results are over a 50\% sample of \dataset{}. }
  \label{tab:cost-summary}
\end{table}

%% file: content/tables/tab_cost_summary.tex
\begin{tabular}{@{}lrrccc@{}}
\toprule
Agent & Cost & Model & New input & P & R \\
 & & calls & + output & & \\
 & & & / corpus & & \\
\midrule
GPT-6-Luna Exa. & \$29.07 & 12,687 & 1.61$\times$ & 43.5 & 80.8 \\\midrule
GPT-6.1-Sol & \$35.58 & 752 & 0.06$\times$ & 64.9 & 62.8 \\
GPT-6-Astra & \$172.90 & 926 & 0.08$\times$ & 65.0 & 69.6 \\
Claude Opus~5.5 & \$317.17 & 15,286 & 2.56$\times$ & 60.9 & 81.9 \\
\bottomrule
\end{tabular}

%% file: content/analysis.tex
\section{Analysis and Discussion}
\label{sec:analysis}

To understand how agents approach exhaustive search, we examine their trajectories and tool outputs.
We observe two primary strategies: writing rule-based heuristics to search the corpus and distributing documents among sub-agents.

\begin{figure*}[t]
  \centering
  \includegraphics[width=0.86\textwidth]{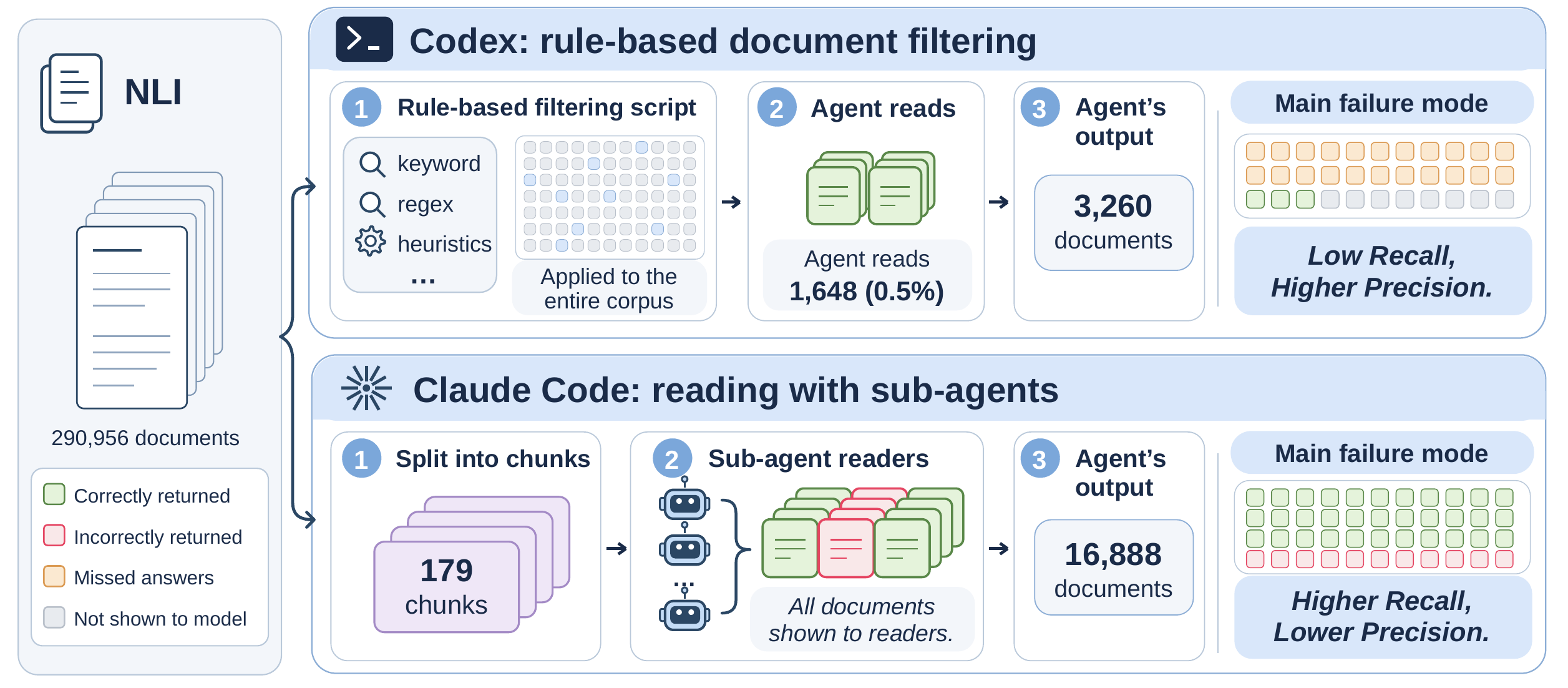}
  \caption{The different search strategies displayed by Codex and Claude. 
  GPT-6.1-Sol uses rule-based search, while Opus~5.5 divides the corpus into 179 chunks for Haiku~5.5 sub-agents to label individually.
  }
  \label{fig:pipeline}
\end{figure*}

\subsection{Strategy 1: Searching with Rules}
\label{sec:strategies}
\label{sec:codex}
\label{sec:workflow}

When agents can no longer read every document directly in context, searching large corpora requires tool use to process thousands of documents.
We observe that an agent inspects a small sample of documents, writes rules in a program to filter answers, apply such rules across the corpus, and iteratively refines the rules after examining sample matches.

Figure~\ref{fig:pipeline} illustrates this workflow on NLI, where the question is to find every claim contradicted by its accompanying passage.
GPT-6.1-Sol first prints 295 documents, and writes rules targeting contradictions through cues in negation, numbers, actions, and dates.
It applies these rules across all 290,956 documents, repeatedly inspecting sample matches and refining the rules over several rounds before returning 3,260 documents as answers (Figure~\ref{fig:traj-nli-rules}).

\paragraph{Low-coverage search.}
\label{sec:shown}
Running a set of filtering rules over the corpus means that the language model is not reading every document, which might be important for semantic understanding and judgment.
To measure how much the model actually reads, we track document text appearing in tool outputs.
We define a document as \textit{never shown} in context if none of its unique five-word sequences appear in tool outputs, and as \textit{mostly shown} if at least half of those sequences appear (Appendix~\ref{app:coverage}).

\begin{figure}[t]
  \centering
  \includegraphics[width=\columnwidth]{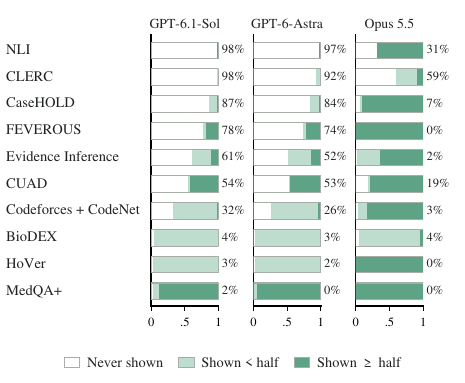}
  \caption{How much of the corpus that was read by the agent.
  The three color bars indicate how much of a benchmark's text appeared in the agent's trajectory. Bold numbers indicate the share of document ``never-shown'' to the agent. Results averaged over ten 50\% samples of \dataset{}.}
  \label{fig:coverage}
  
\end{figure}

As shown in Figure~\ref{fig:coverage}, agents read very little of the corpus in their context.
On GPT-6.1-Sol's 100\% corpus NLI run, 287,809 of the 290,956 documents are \textit{never shown} (98.92), and 1,523 of 2,057 correct answers it found were \textit{never shown}.
So despite not reading most of the corpus, agents can still find a lot of answers using rules.
In addition, we find that as corpus size increases, the number of \textit{mostly shown} documents drop, so agents don't read more.
For GPT-6.1-Sol on NLI, the number of \textit{mostly shown} documents drops from 4,510 at 25\% corpus to 1,338 at 100\% corpus.
This helps explain why recall falls, since agents rely even more on pre-defined rules to scan larger corpora, missing many answers that these rules leave out.
In particular, for Codex on NLI, CLERC, and CaseHOLD, 84--98\% of missed answers are \textit{never shown} in context. 
In contrast, when an answer document is \textit{mostly shown}, it is returned 91\% of the time.

\paragraph{Why rules miss answers.}
The filtering rules which agents define often fail to capture every answer document.
For example, on Codeforces + CodeNet, where the goal is to find rejected submissions, GPT-6.1-Sol runs submitted solutions against the provided sample inputs, and then investigates their failures.
Of the 305 judge-rejected submissions, 221 (72\%) pass all sample tests and fail only on hidden tests, so the agent's tests cannot flag them. 
As a result, it misses 197 of those 221 answers (Figure~\ref{fig:traj-code-rules}).

FEVEROUS illustrates a different limitation, where the lexical heuristics came up by agents miss genuine contradictions. 
For example, GPT-6-Astra filters claims by checking word overlap between claims and evidence. 
In one case, a claim swaps two numerical ratings between reviewers; because both ratings still appear in the evidence, the word-matching rule filters out the document. 
Overall, 269 of the 404 missed answers in this run never reach the model (Figure~\ref{fig:traj-feverous}).

\paragraph{Lack of acknowledgment.}
\label{sec:silent}
Despite leaving documents unread, Codex agents rarely report that their search was incomplete (Appendix~\ref{app:stopping}). 
Across 200 Codex runs on the 50\% corpora, 184 never acknowledge in their final message that the search was not exhaustive.
Opus~5.5 admits in its final message that the results may be incomplete in 72 of 100 runs and concedes only a limitation in 27 others.

\subsection{Strategy 2: Reading with Sub-agents}
\label{sec:claudecode}

Beyond rules, agents also show the behavior of reading documents.
For example, Claude Code natively supports sub-agents, allowing Opus~5.5 to partition a corpus into batches and assign each batch to a sub-agent for review.
It does so in 99 of its 100 runs, spawning a median of 56.5 sub-agents per run.
Figure~\ref{fig:pipeline} illustrates this workflow on NLI: Opus~5.5 divides the 290,956 documents into 179 chunks and asks Haiku~5.5 sub-agents to evaluate every document, raising recall from 17.7 to 79.7.

\paragraph{Errors in multi-agent workflow.}
Even though more documents are read, poor aggregation and coordination can introduce many false positives.
In the setting of Figure~\ref{fig:pipeline}, the main Opus~5.5 orchestrator agent prompts its sub-agents to flag documents as "contradictions", "uncertain," or "irrelevant." 
Among the 5,161 documents returned as "uncertain," Opus~5.5 inspects a sample of 30 documents and subsequently chooses to include all "uncertain" documents in the answer set, leading to a recall of 79.7 but only 54.9 precision.
Furthermore, on MedQA+, where agents must identify clinical notes containing medical errors, Opus~5.5 prompts its sub-agents to expect errors in about 50\% of documents, even though only 3.8\% actually contain them.
Sub-agents then spends two rounds of review to flag 7.4\% of notes, reaching 96.7 recall but only 50.6 precision (Figure~\ref{fig:traj-medqa}).
The low precision is caused often by genuine agent errors rather than dataset errors (Table~\ref{tab:label-audit-agents}).

\paragraph{Falling back to rules.}
Opus~5.5 does not read every document on all tasks.
While at least 80\% of documents are \textit{mostly shown} on some tasks, Opus~5.5 relies on heuristic rules on CLERC, Evidence Inference, and BioDEX.
On CLERC and BioDEX, it first filters the corpus with keyword-based rules and sends only matching excerpts to its sub-agent readers. 
On Evidence Inference, it keeps every record, but its readers mostly see short excerpts of each article, chosen by keyword overlap with the record's outcome.
As an example, most missed answers on CLERC never reach the model (Figure~\ref{fig:coverage}).
On BioDEX, Opus~5.5 sends its sub-agents only excerpts of the articles that pass its keyword screen and asks the sub-agent to rate how likely an article is to be an answer.
It then keeps articles rated above a cutoff it chooses, and most missed answers (196 of 232) pass the screen but are rated below it by the sub-agent (Figure~\ref{fig:traj-biodex}).

\paragraph{Summary of trade-offs.}
Taken together, these two strategies reveal two distinct challenges in exhaustive search.
Searching using rules is efficient, but often fails because the rules do not reliably find all answer documents.
Reading with sub-agents gives coverage, but still requires the agent to combine its sub-agents outputs reliably, and is prohibitively expensive.

%% file: content/related_work.tex
\section{Related Work}

% \tomer{Think about making this a "Background and Related Work" section to contrast what's been done before}

\paragraph{Agentic search benchmarks.}
Many search benchmarks test whether an agent can find a single hard-to-find answer~\citep{wei2025browsecomp,chen2025browsecomp}.
More recent work asks for complete answer sets, both in web search, where agents often stop early~\citep{jang2026odysseys,vinogradov2026hunt}, and in multi-document question answering~\citep{amouyal2022qampari,malaviya2023quest,wolfson2026monaco,gupta2026deepsearchqa,wong2026widesearch,rafiee2026total, thakur2026orbitscalableverifiabledata,chang2026karlknowledgeagentsreinforcement}.
Closest to our work are search tasks that require document classification~\citep{ravfogel2024description,patel2025semanticoperatorsdeclarativemodel,chen2025enrichindex,shankar2025docetl,wei2026moar,zhou2026satir}, and in particular OBLIQ-Bench, which shows that models can verify relevant documents that retrievers miss~\citep{tchuindjo2026obliq}.
Unlike OBLIQ-Bench, which tests top-$k$ retrieval on partially labeled corpora, \dataset{} fully labels closed collections, letting us scale corpus size and measure exact recall, coverage, and cost.

\paragraph{Long-context reasoning.}
Work on reasoning over large corpora has taken two main directions.
One puts the full corpus into a single prompt, where benchmarks reveal failures in multi-answer retrieval, aggregation, and spotting missing items~\citep{hsieh2024ruler,bertsch2025oolong,fu2026absence,lee2024can,vodrahalli2024michelangelo,zeng2026locabench}.
In this setting, CTC-Bench shows that efficient attention mainly degrades when tasks compare across documents rather than judge each document alone~\citep{singhal2026no}.
A second direction moves beyond one prompt using coding agents and recursive language models that run code, read files, and invoke sub-models~\citep{cao2026coding,zhang2025recursive}.
Our findings qualify this agentic approach: even when each document can be judged on its own, frontier agents still fail to search large corpora completely and efficiently.

%% file: content/conclusion.tex
\section{Conclusion}

Our findings underscore the importance of evaluating exhaustive search as a distinct capability of LLM-based agents, since finding correct answers does not necessarily mean that a search is complete. Even frontier agents exhibit notable limitations but in different ways: Codex-based agents often fall short of exhaustive coverage, whereas Claude Code can be substantially less efficient than a simple exhaustive-search baseline. These findings highlight the need for further research into making agentic exhaustive search more effective and efficient. In particular, search agents should be able to recognize when their search remains incomplete and adapt their strategies to the characteristics of the underlying corpus. We hope that \dataset{} provides a valuable testbed for investigating these challenges and helping the development of more accurate and efficient agents for exhaustive search.

\section*{Limitations}

The datasets that comprise \dataset{} (\S\ref{sec:benchmark}) rely on annotations from existing datasets, which may contain errors or omit valid answers, causing us to underestimate precision. Our analysis in Appendix~\ref{app:experiments:error_analysis} estimates the prevalence of dataset errors by analyzing agent's true positive results. Nevertheless, since all agents are evaluated over the same data their relative ranking in terms of precision should remain the same.
Furthermore, we use the same answer keys across corpus sizes, so these limitations are unlikely to explain the observed decline in recall.

Our experiments center on three frontier agents which use the Codex and Claude Code harnesses (\S\ref{sec:evaluation}).
In our coverage analysis (\S\ref{sec:analysis} we measured which documents are provided as input to an agent's context using n-gram overlap. This method may somewhat overestimate how much the agent actually reads.

%% file: content/appendix/main.tex
\clearpage
\appendix
\makeatletter
\setlength{\@dblfptop}{0pt}
\setlength{\@dblfpsep}{16pt}
\setlength{\@dblfpbot}{0pt plus 1fil}
\makeatother
\input{content/appendix/tasks}

\input{content/appendix/questions}

\input{content/appendix/agents}

\input{content/appendix/results}
\clearpage
\input{content/appendix/coverage}
\input{content/appendix/workflows}
\input{content/appendix/trajectories}

%% file: content/appendix/tasks.tex
\section{Task Constructions}
\label{app:tasks}

We describe the original datasets and how we adapt each into an exhaustive-search task, including the construction of its corpus, questions, and answer key. 
Token counts use the o200k tokenizer~\citep{openai2024tiktoken}.

For each dataset, we first build the largest possible corpus, and then subsample from it to create different copora while keeping the question fixed.
Tables~\ref{tab:q1}--\ref{tab:q2c1} give every task question verbatim, with one example answer each.
Tables~\ref{tab:prompts} and~\ref{tab:answer-formats} give the prompts and answer formats.

\subsection{SNLI,  MultiNLI, ANLI and VitaminC}\label{app:task-nli}
We combine four natural language inference (NLI) and fact-verification benchmarks: SNLI~\citep{bowman-etal-2015-large}, MultiNLI~\citep{williams-etal-2018-broad}, ANLI~\citep{nie2020adversarial} and VitaminC~\citep{schuster2021get}.
SNLI, MultiNLI, and ANLI respectively comprise 570,152, 412,702, and 169,265 (hypothesis, premise) pairs each labeled as \emph{entailment}, \emph{contradiction}, or \emph{neutral}.
VitaminC is a fact-verification dataset with 488,904 (claim, evidence) pairs labeled as \emph{supports}, \emph{refutes}, or \emph{not enough info}.
We build one corpus from these pairs and ask for all contradiction pairs (Table~\ref{tab:q1}), that is, the \emph{contradiction} class in ANLI, SNLI, and MultiNLI, and the \emph{refutes} class in VitaminC.

\paragraph{ANLI.} 
We remove 166 duplicate pairs and 4 pairs with inconsistent labels. This gives 169,095 pairs, of which 44,052 are \emph{contradictions}.

\paragraph{VitaminC.} 
We first remove the 163,180 synthetic pairs.
For the remaining 325,724 pairs from 106,954 Wikipedia edits, we choose one pair per edit randomly, giving 106,954 pairs.
We then remove 158 duplicates pairs, giving a corpus of 106,796 pairs, of which 23,356 are \emph{refutes}.

\paragraph{SNLI and MultiNLI.} 
We remove duplicate pairs (655 from SNLI and 167 from MultiNLI) as well as pairs lacking a gold label (1,119 from SNLI and 353 from MultiNLI). This yields 568,378 SNLI pairs (189,588 contradictions) and 412,182 MultiNLI pairs (137,316 contradictions).

\paragraph{Final Corpus.}
We randomly keep one claim per unique passage, leaving 10,760 ANLI, 98,211 VitaminC, 157,212 SNLI and 134,488 MultiNLI passages.
Then we randomly sample contradictions to make up 4\% of documents.
The full corpus contains 290,956 documents (15.5M tokens), of which 11,638 are answers: 7,999 ANLI documents (320 answers), 80,648 VitaminC (3,226), 109,002 SNLI (4,360) and 93,307 MultiNLI (3,732).

\subsection{HoVer}\label{app:task-hover}
HoVer~\citep{jiang2020hover} is a fact-verification benchmark with 26,171 claims, each of which has 2 to 4 Wikipedia articles as evidence.
Annotators labeled each claim as \emph{supported} or \emph{not supported}.
We ask for all claims that their articles are in the \emph{not supported} class (Table~\ref{tab:q1}).

\paragraph{Documents.} 
The train and dev splits contain 22,171 claims (13,023 supported and 9,148 unsupported) derived from 7,624 source questions.
We leave out the test split because it does not have publicly available labels.
Because HoVer creates multiple closely related variants per question, we keep one claim per question by selecting randomly. We keep 7,230 available supported claim.

Each claim is then paired with 7 to 9 Wikipedia articles, including 2 to 4 annotated gold evidence articles and 5 distractor articles taken from HoVer's top retrieval results. We then shuffle these paragraphs into a random order.

\paragraph{Final Corpus.}
We take not-supported claims as the answers. We randomly sample not-supported claims to make up 4\% of our corpus, giving 290 claims.
For 272 of such claims, the question is already in our supported set, so in this case the non-supported claim replaces the supported claim. 
We then take the other 18 non-supported claims from 18 questions outside of our supported set, giving a corpus 6,958 supported and 290 not-supported claims. The final corpus contains 7,248 documents (7.0M tokens) with 290 answers.

\subsection{FEVEROUS}\label{app:task-feverous}
FEVEROUS~\citep{aly2021feverous} is a fact-verification benchmark with 87,026 claims about Wikipedia.
Each claim is labeled as \emph{supports}, \emph{refutes}, or \emph{not enough info}, and comes with an annotated evidence set including passages, tables, and lists.
We ask for all claims in the \emph{refutes} class (Table~\ref{tab:q1}).

\paragraph{Documents.}
The train and dev splits contain 79,181 claims (45,743 supported, 30,696 refuted, and 2,742 with not enough info).
We leave out the test split because it does not have publicly available labels.
Each claim is paired with its Wikipedia evidence set.

\paragraph{Final Corpus.}
We take refuted claims as the answers.
We keep all other claims, and randomly sample refuted claims to make up 4\% of our corpus, giving 2,020 claims.
The final corpus contains 50,505 documents (7.2M tokens) with 2,020 answers.

\subsection{Codeforces + CodeNet}
\label{app:task-code}

The Codeforces collection~\citep{penedo2025codeforces} contains 12.6 million submissions to 10,024 programming problems, and CodeNet~\citep{puri2021codenet} contains 13.9 million submissions to 4,053 problems from the AIZU and AtCoder judges.
Each submission is labeled with the verdict of the online judge, such as \emph{accepted}, \emph{wrong answer}, \emph{runtime error}, or \emph{time limit exceeded}.
Our corpus contains only submissions with three verdicts, \emph{accepted}, \emph{wrong answer}, and \emph{runtime error}, and we ask for all submissions with \emph{wrong answer} or \emph{runtime error} (Table~\ref{tab:q1b}).
This is because the results of other verdict categories often change depending on the hardware used (e.g. \emph{time limit exceeded}).

\paragraph{Codeforces.}
After removing interactive problems (which need multiple rounds of feedback) and duplicates, we keep 9,734 problems and their 1,081,179 valid Python 3 submissions.
We then remove 131,339 rejected submissions that fails the sample input, since they are very easy to judge.

\paragraph{CodeNet.}
We keep the 3,888 problems with a complete problem description and specified time/space limits (2,369 from AIZU and 1,519 from AtCoder), with their Python 3 and C++ submissions.
We then removed 2,336 AIZU problems because the judge can be inconsistent with problem statements.

\paragraph{Final Corpus.}
We remove the submissions rejected for exceeding the time limit (362) or the memory limit (344). 
We keep every accepted submission (5,804) as a non-answer and sample \emph{wrong answer} and \emph{runtime error} submissions as the answers, to 5\% of documents.
The final corpus contains 6,109 documents (5.1M tokens) with 305 answers: 161 \emph{wrong answer} and 144 \emph{runtime error}.

\subsection{CaseHOLD}
\label{app:task-casehold}

CaseHOLD~\citep{zheng2021does} is a legal reasoning benchmark comprising 53,137 passages from U.S. court opinions. 
Each passage cites a prior case with its holding masked out, accompanied by five candidate holdings (one correct and four incorrect). 
To adapt it for an exhaustive-search task, we insert one candidate holding into each passage and ask for all passages containing an incorrect holding (Table~\ref{tab:q1b2}).

\paragraph{Documents.}
We exclude 575 passages that do not contain exactly one holding placeholder (573 with none and 2 with two) along with 2 duplicate passages. 
Each document consists of a single passage with one candidate holding inserted and enclosed in double square brackets.

\paragraph{Final Corpus.}
For a random 4\% of passages, we insert one of CaseHOLD's four incorrect candidates at random, forming the answer set. 
For the remaining passages, we insert the correct candidate. 
Since CaseHOLD draws its incorrect candidates from the correct holdings of other passages, an inserted incorrect holding would also appear, word for word, as the correct holding of another passage.
Therefore, to avoid the same holding appearing twice, we remove the 1,981 passages whose correct holding is inserted as an incorrect candidate elsewhere.
The final corpus contains 50,579 documents (12.7M tokens) with 2,102 answers.

\subsection{CLERC}
\label{app:task-clerc}

CLERC~\citep{hou2025clerc} is a legal case retrieval benchmark constructed from 1.84M U.S. federal court decisions. 
Each query consists of a 300-word passage surrounding a citation that is removed, and the task is to retrieve the cited case. 
We adapt it by asking for all passages that cite one of the following Supreme Court decisions: \emph{Terry v. Ohio}, \emph{Illinois v. Gates}, \emph{United States v. Leon}, and \emph{Schneckloth v. Bustamonte} (Table~\ref{tab:q1c1}).

\paragraph{Cases.}
Besides the four target cases, we pick 73 other cases on Fourth Amendment, such as other decisions on when police may stop a person (\emph{United States v. Sokolow}, \emph{Illinois v. Wardlow}). 

\paragraph{Documents.}
We search all 1.84M decisions for citations of the 77 cases (e.g., ``392 U.S. 1'' for \emph{Terry}) and, following CLERC, extract the 300 words centered on each citation. 
From each decision, we keep at most one passage per cited case, yielding 61,431 passages. 
A passage's label is the case cited at its center.
We then remove 49 passages that are from the four target opinions themselves and 441 duplicate passages.

\paragraph{Masking.}
CLERC's single-removal setting removes only the central citation and keeps all other citations visible. 
We start from this setting, but replace the central citation in place with [CITATION] instead of deleting its sentence, and also mask every other reference to the same case in the passage: its short forms, parallel citations, pin cites, and ``Id.'' become [CITATION], and its name becomes [CASE]. 
Citations to all other cases stay untouched and visible. 
We mask every passage this way, whether its central case is a target or one of the 73 other cases. 
We then remove 806 passages where the case could not be fully masked, mostly citations of recent decisions printed without a page number.
Finally, we add the full opinions of the four target cases as reference documents, which belong to no answer.

\paragraph{Final Corpus.}
The final corpus comprises 60,139 documents (25.3M tokens), consisting of 60,135 passages and the four reference opinions. 
Across the four questions, the answer counts are 5,063 for \emph{Terry} (8.4\%), 3,156 for \emph{Gates} (5.2\%), 2,244 for \emph{Leon} (3.7\%), and 2,397 for \emph{Schneckloth} (4.0\%). 
The remaining 47,275 passages cite the 73 other cases.

\subsection{CUAD}
\label{app:task-cuad}

CUAD~\citep{hendrycks2021cuad} is a contract-review benchmark of 510 legal contracts in which lawyers annotated 13,101 contract clauses with 41 relevant categories. 
We follow LegalBench's~\citep{guha2023legalbench} adaption of CUAD and treat each highlighted clause as a document while maintaining the labeled category to which it belongs. 
Using CUAD's original category descriptions we formulate four questions, each asking for all clauses belonging to a specific target category: (1) Exclusivity, (2) Post-Termination Services, (3) Revenue/Profit Sharing, and (4) Governing Law. The task prompt is presented in Table~\ref{tab:q1c1}. 
The documents associated with every single category are fixed at approximately 5\% of the corpus.

\paragraph{Documents.}
Each document corresponds to a single clause as listed in CUAD.
When a clause spans non-contiguous passages in a contract, we retain CUAD's ``<omitted>'' marker between them. 
We de-duplicate 2,057 clauses with identical text by merging them into a single document that inherits all of their category labels. 
We then remove 2,706 contract clauses shorter than eight words, which consist mostly of party names, dates, and document titles, and also remove 32 clauses that are under-specified.

\paragraph{Final Corpus.}
For each question, the answers include all clauses labeled with that category in CUAD, along with 36 overlapping clauses from the same contracts where the text was annotated multiple times.
The final corpus contains 8,306 documents (0.68M tokens), with 410 answers for Exclusivity (4.9\%), 419 for Post-Termination Services (5.0\%), 394 for Revenue/Profit Sharing (4.7\%), and 440 for Governing Law (5.3\%).

\subsection{Evidence Inference}
\label{app:task-evinf}

Evidence Inference 2.0~\citep{deyoung2020evidence} is a benchmark of articles reporting randomized controlled trials, from PubMed Central. 
For each article, medical doctors wrote comparisons—each specifying an intervention (e.g. aspirin), a comparator (e.g. a placebo), and an outcome (e.g. duration of headache)—and labeled the finding as \emph{significantly increased}, \emph{significantly decreased}, or \emph{no significant difference} based on the full text. This label indicates whether the measurable outcome went down (decreased) in the treatment group relative to the comparator group. 
In \dataset{}, we ask the agent for every comparison whose result is \emph{significantly decreased} (Table~\ref{tab:q2}).

\paragraph{Documents.}
Each document consists of a single comparison paired with the full text of its article.
Several comparisons may have the same article (8,162 pairs with 2,898 distinct articles). 
Starting from the 12,578 comparisons with a valid label, we exclude 21 with conflicting doctor labels and 153 whose article text is empty or shorter than 1,000 tokens. 
We also remove 87 comparisons flagged by the original authors as incorrect, questionable, or malformed,\footnote{\url{https://github.com/jayded/evidence-inference/blob/master/annotations/README.md}} as well as 1,640 comparisons annotated by only a single doctor instead of two.

\paragraph{Final Corpus.}
We define the answers as comparisons labeled \emph{decreased}. 
We include all the non-target comparisons (\emph{increased}, \emph{no difference}) and down sample the target comparisons (\emph{decreased}) to be 4.7\% of the corpus. 
The final corpus contains 8,162 documents (52.3M tokens) and 383 target documents (4.7\% answers).

\subsection{MedQA+}
\label{app:task-medqa}

MEDEC~\citep{abacha2025medec} is a benchmark for detecting medical errors in clinical texts. 
Its MS collection is built from MedQA exam questions by removing the question from each patient case and inserting a single sentence: either the correct answer or a wrong option. 
MedQA+ applies this same construction to three public exam banks: MedQA~\citep{jin2021disease}, MedMCQA~\citep{pal2022medmcqa}, and MMLU professional medicine~\citep{hendrycks2020measuring}. 
We merge MedQA+ with MEDEC and use MEDEC's original prompt to ask for every clinical text that contains a medical error (Table~\ref{tab:q2c1}). 

\paragraph{Documents.} 

MEDEC-MS contains 3,360 clinical texts, each labeled as correct or as containing an error.
Most unique patient cases appear twice, once with the correct answer and once with a wrong one.
For the 1,497 cases with both versions, we retain one text per case, selecting the error version in a random 5\% of cases.

MedQA+ adapts MEDEC's construction to the three other medical exam benchmarks mentioned above using a regex script. 
We exclude image-dependent questions, negated questions (e.g. ``all of the following except''), questions with options such as ``all of the above'', and near-duplicates. 
For the remaining questions, the script strips the question prompt (e.g., ``Which of the following is the most likely diagnosis?'') and inserts a declarative answer sentence at a sentence boundary, usually the end, so the case reads as a clinical note. 
Each sentence is sampled from five to eight fixed templates matched to the question type (e.g., ``The patient is diagnosed with \emph{answer}.''\ or ``\emph{Answer} is the causal organism.''). 
In 4\% of notes per exam benchmark, the script instead substitutes a wrong option to create an error note. 
This pipeline yields 7,876 clinical notes.

To further ensure label quality, we launched Claude Sonnet~5.5 to audit every inserted sentence individually.
For each note, the model either makes no changes, or rewrites it to answer the original question (495 notes), or drops the note when the question is still not answerable (e.g. requires external content such as radiograph, 545 notes). 

\paragraph{Final Corpus.}
We retain only texts whose inserted answer is a diagnosis, a cause, or an organism, leaving 531 MEDEC texts and 4,600 MedQA+ notes. 
This removes 966 MEDEC texts (659 management, 154 treatment, 149 pharmacotherapy, 4 unspecified) and 2,731 MedQA+ notes (1,973 management, 496 treatment, 262 drug), because these categories are more open-ended.
We then remove texts that cannot be judged or whose label is unreliable (48 MEDEC texts that refer to a missing figure; 24 MedQA+ notes whose answer key is unspecified), and 383 MedQA+ notes that repeat a MEDEC case.
The final corpus comprises 4,676 documents (0.56M tokens) with 189 answers (4.0\%), including 483 MEDEC texts (24 errors) and 4,193 MedQA+ notes (165 errors).

\subsection{BioDEX}
\label{app:task-biodex}
BioDEX~\citep{d2023biodex} is a benchmark for extracting adverse drug events from biomedical articles. 
Each article (title, abstract, and text) is paired with a safety report, filed with the FDA by drug-safety staff after reading the article, that lists the patient's drugs and reactions using MedDRA terms.
We ask the agent 5 questions, each associated with a different adverse reaction: acute kidney injury, neutropenia, thrombocytopenia, pyrexia, and respiratory failure.
The agent is tasked with finding every article in which that reaction was part of an adverse drug event (Table~\ref{tab:q2c1}).

\paragraph{Documents.}
Each document is a single article from BioDEX. 
Of the 15,659 available articles,\footnote{https://huggingface.co/datasets/BioDEX/BioDEX-ICSR} we removed 5 that have missing text.

\paragraph{Final Corpus.}
An article is an answer to a question if its safety report lists that reaction.
This does not guarantee that the reaction itself appears verbatim in the article text, e.g. using the medical term \emph{Pyrexia} instead of fever. Second, the reaction can be present in an article while not being the adverse drug event, e.g. ``thrombocytopenia'' is a reaction, but may also appear as a patient’s existing condition, before taking any drug.
We keep all 1,559 articles that answer at least one question, and we randomly sample from the remaining articles so that the five target reactions make up around 5\% of the corpus. 
The final corpus contains 7,108 documents (25.7M tokens): 463 answers for acute kidney injury (6.5\%), 360 for neutropenia (5.1\%), 349 for thrombocytopenia (4.9\%), 312 for pyrexia (4.4\%), and 293 for respiratory failure (4.1\%).

%% file: content/appendix/questions.tex
\begin{table*}[p]
\centering\scriptsize
\setlength{\tabcolsep}{5pt}
\renewcommand{\arraystretch}{1.08}
\begin{tabularx}{\textwidth}{@{}>{\raggedright\arraybackslash\itshape}p{0.105\textwidth}>{\raggedright\arraybackslash}X@{}}
\toprule
\multicolumn{2}{@{}l}{\textbf{NLI}} \\
Message & Read question.txt in this directory and answer it. The corpus is the pages/ directory. Write your final answer to answer.txt as a JSON array of record ids. \\
\cmidrule(l{0pt}r{0pt}){1-2}
Question & Each record holds a statement and a passage. The passage is a short text: a paragraph from Wikipedia or another source, the caption of a photo, or a line from a text. When the passage describes a scene, a situation or an event, the statement is about the same scene, situation or event. Using the passage and what you know about the world, a statement is definitely correct (the passage supports it), definitely incorrect (the passage refutes it), or neither (the passage does not contain enough information; the statement might be correct). Find every record whose statement is definitely incorrect given its passage. The passage is the source of truth: a statement that an out-{}of-{}date passage refutes counts, even if the statement is true today. Decide each record from its own passage only.\par
Write your final answer to answer.txt as a JSON array of record ids, exactly as each record\textquotesingle{}s first line gives them (for example \textquotedbl{}m0a1b2c3d4\textquotedbl{}). \\
\cmidrule(l{0pt}r{0pt}){1-2}
Example answer & \emph{Record m7a882aab0 (VitaminC).} Statement: Mauritania is officially referred to as the People\textquotesingle{}s Republic of Mauritania. \ldots{} Mauritania (;; or;; Soninke: Murutaane; Pulaar: Moritani;), officially the Islamic Republic of Mauritania, is a country in the Maghreb region of Northwestern Africa. \\
\cmidrule(l{0pt}r{0pt}){1-2}
Why it qualifies & The passage gives the official name as the Islamic Republic of Mauritania, not the People's Republic. \\
\midrule
\multicolumn{2}{@{}l}{\textbf{HoVer}} \\
Message & Read question.txt in this directory and answer it. The corpus is the pages/ directory. Write your final answer to answer.txt as a JSON array of claim labels. \\
\cmidrule(l{0pt}r{0pt}){1-2}
Question & Each file in pages/ holds one claim, labeled like \textquotedbl{}Claim 4821\textquotedbl{}, followed by the opening paragraphs of several Wikipedia articles (as of 2017) collected for that claim. Some of the articles bear on the claim and some do not.\par
Question. Find every claim that is not supported by the articles in its file. A claim is supported if, without using external knowledge, you have strong reason to believe the claim is true given the articles. A claim is not supported if it is refuted (without using external knowledge, it is impossible for the claim to be true given the articles) or if there is not enough information (you need more information other than the given articles to say whether the claim is true or false). Your knowledge of English (for example, being a captain entails also being a leader) and common sense (for example, a person can only have one birth place) are not external knowledge. Decide each claim from its own file only.\par
Write your final answer to answer.txt as a JSON array of the claim labels, exactly as written in the files (for example [\textquotedbl{}Claim 4821\textquotedbl{}, \textquotedbl{}Claim 77\textquotedbl{}]). \\
\cmidrule(l{0pt}r{0pt}){1-2}
Example answer & \emph{Claim 452277; a 345-{}word file with eight articles.} Neither the magazine Ángel Manuel Rodríguez authors in and Girls\textquotesingle{} Life have a headquarter or editor based in Maryland.\newline Articles:\newline [Girls\textquotesingle{} Life (magazine)]\newline Girls\textquotesingle{} Life is an American teen magazine. It was launched in 1994 by Monarch Services. The magazine is headquartered in Baltimore, Maryland. \ldots{} [Ángel Manuel Rodríguez]\newline Ángel Manuel Rodríguez (1945—) is a Seventh-{}day Adventist theologian and was the director of the Biblical Research Institute (BRI) before his retirement. His special research interests include Old Testament, Sanctuary and Atonement, and Old Testament Theology. He has written several books, and authors a monthly column in \textquotedbl{}Adventist World\textquotedbl{}.\newline [Adventist World]\newline Adventist World is a monthly international magazine of the Seventh-{}day Adventist Church published by the Review and Herald Publishing Association. Editors are based in Silver Spring, Maryland and Seoul, Korea. \\
\cmidrule(l{0pt}r{0pt}){1-2}
Why it qualifies & Girls' Life is headquartered in Baltimore, Maryland, and Adventist World's editors are in Silver Spring, Maryland. \\
\midrule
\multicolumn{2}{@{}l}{\textbf{FEVEROUS}} \\
Message & Read question.txt in this directory and answer it. The corpus is the pages/ directory. Write your final answer to answer.txt as a JSON array of claim labels. \\
\cmidrule(l{0pt}r{0pt}){1-2}
Question & Each file in pages/ holds one claim, labeled like \textquotedbl{}Claim 4821\textquotedbl{}, followed by the evidence collected for it from Wikipedia (as of December 2020): sentences, table cells and list items, under the title of their article in square brackets and the section they come from. A table is shown as the selected cells and their header cells, one line per table row, cells separated by \textquotedbl{} $|$ \textquotedbl{}; a cell that spans several rows or columns is repeated, and \textquotedbl{}...\textquotedbl{} stands for a cell that is not part of the evidence.\par
Question. Find every claim that is refuted by the evidence in its file. The evidence of a claim supports it, refutes it, or does not provide enough information to reach a verdict. As a guide, ask yourself: if I was given only the selected sentences, table cells, and list items, do I have strong enough reason to believe the claim is supported or strong enough reason to believe the claim is refuted? For supported claims, every piece of information has to be verified by evidence, whereas for refuted claims, the evidence only needs to be sufficient to refute one part of the claim. The evidence does not have to contradict the claim directly: would you consider yourself misled by the claim given the evidence? If yes, the claim is refuted (for example, the claim \textquotedbl{}Shakira is Canadian\textquotedbl{} is refuted by the evidence \textquotedbl{}Shakira is a Colombian singer, songwriter, dancer, and record producer\textquotedbl{}). You must apply common-{}sense reasoning to the evidence you read but avoid applying your own (world) knowledge. Decide each claim from its own file only.\par
Write your final answer to answer.txt as a JSON array of the claim labels, exactly as written in the files (for example [\textquotedbl{}Claim 4821\textquotedbl{}, \textquotedbl{}Claim 77\textquotedbl{}]). \\
\cmidrule(l{0pt}r{0pt}){1-2}
Example answer & \emph{Claim 6691.} Cuspidia is an animal from the Asterales order.\newline Evidence:\newline [Cuspidia]\newline Cuspidia is a genus of flowering plants in the daisy family. \\
\cmidrule(l{0pt}r{0pt}){1-2}
Why it qualifies & The evidence calls Cuspidia a genus of flowering plants, not an animal. \\
\bottomrule
\end{tabularx}
\caption{Agent prompts and example answers (target documents) in the fact-checking domain.}
\label{tab:q1}
\end{table*}

\begin{table*}[p]
\centering\scriptsize
\setlength{\tabcolsep}{5pt}
\renewcommand{\arraystretch}{1.08}
\begin{tabularx}{\textwidth}{@{}>{\raggedright\arraybackslash\itshape}p{0.105\textwidth}>{\raggedright\arraybackslash}X@{}}
\toprule
\multicolumn{2}{@{}l}{\textbf{Codeforces + CodeNet}} \\
Message & Read question.txt in this directory and answer it. The corpus is the pages/ directory. Write your final answer to answer.txt as a JSON array of record ids. \\
\cmidrule(l{0pt}r{0pt}){1-2}
Question & Each record in this collection is a programming problem from an online judge (its statement, time and memory limits, input and output format, and usually examples and notes) followed by one submission to it, headed \textquotedbl{}Solution\textquotedbl{} with the language the submission was judged in; its first line gives the record id.\par
The online judge ran each submission on its own tests for the problem and gave it a verdict. Find every record whose submission the judge did not accept: its verdict is Wrong Answer (the program produced output different from the judge\textquotesingle{}s data) or Runtime Error (the program failed during execution, for example a stack overflow or a division by zero). Every other submission in this collection was accepted: it passed all of the judge\textquotesingle{}s tests. No submission here was rejected for exceeding the time limit or the memory limit. The judge\textquotesingle{}s tests are not given: decide from the record which verdict the judge gave.\par
Write your final answer to answer.txt as a JSON array of record ids, exactly as each record\textquotesingle{}s first line gives them (for example \textquotedbl{}r0a1b2c3d4\textquotedbl{}). \\
\cmidrule(l{0pt}r{0pt}){1-2}
Example answer & Problem: Seating On Bus\newline Time limit: 1 seconds. Memory limit: 256 megabytes.\newline Consider 2n rows of the seats in a bus. n rows of the seats on the left and n rows of the seats on the right. Each row can be filled by two people. So the total capacity of the bus is 4n.\newline Consider that m (m $\leq$ 4n) people occupy the seats in the bus. The passengers entering the bus are numbered from 1 to m (in the order of their entering the bus). The pattern of the seat occupation is as below:\newline 1-st row left window seat, 1-st row right window seat, 2-nd row left window seat, 2-nd row right window seat, ... , n-th row left window seat, n-th row right window seat.\newline After occupying all the window seats (for m $>$ 2n) the non-window seats are occupied:\newline 1-st row left non-window seat, 1-st row right non-window seat, ... , n-th row left non-window seat, n-th row right non-window seat.\newline All the passengers go to a single final destination. In the final destination, the passengers get off in the given order.\newline 1-st row left non-window seat, 1-st row left window seat, 1-st row right non-window seat, 1-st row right window seat, ... , n-th row left non-window seat, n-th row left window seat, n-th row right non-window seat, n-th row right window seat.\newline The seating for n = 9 and m = 36.\newline You are given the values n and m. Output m numbers from 1 to m, the order in which the passengers will get off the bus.\newline Input\newline The only line contains two integers, n and m (1 $\leq$ n $\leq$ 100, 1 $\leq$ m $\leq$ 4n) --- the number of pairs of rows and the number of passengers.\newline Output\newline Print m distinct integers from 1 to m --- the order in which the passengers will get off the bus.\newline Examples\newline Input:\newline 2 7\newline Output:\newline 5 1 6 2 7 3 4\newline Input:\newline 9 36\newline Output:\newline 19 1 20 2 21 3 22 4 23 5 24 6 25 7 26 8 27 9 28 10 29 11 30 12 31 13 32 14 33 15 34 16 35 17 36 18\newline Solution (Python 3, CPython): \newline \texttt{n,m = map(int,input().split())} \newline \texttt{s = [[i for i in range(j,4*n+1,2)] for j in range(1,n+1)]} \newline \texttt{r = [s[0][0:n],s[0][n::]]} \newline \texttt{l = [s[1][n::],s[1][0:n]]}\newline \texttt{a = []}\newline \texttt{for i in range(n):}\newline \texttt{\mbox{}\ \ \ \ if r[1][i]<=m:}\newline \texttt{\mbox{}\ \ \ \ \ \ \ \ a.append(r[1][i])}\newline \texttt{\mbox{}\ \ \ \ if r[0][i]<=m:}\newline \texttt{\mbox{}\ \ \ \ \ \ \ \ a.append(r[0][i])}\newline \texttt{\mbox{}\ \ \ \ if l[0][i]<=m:}\newline \texttt{\mbox{}\ \ \ \ \ \ \ \ a.append(l[0][i])}\newline \texttt{\mbox{}\ \ \ \ if l[1][i]<=m:}\newline \texttt{\mbox{}\ \ \ \ \ \ \ \ a.append(l[1][i])}\newline \texttt{print(*a)} \\
\cmidrule(l{0pt}r{0pt}){1-2}
Why it qualifies & For $n=1$, \texttt{s} has one row, so \texttt{s[1]} raises an \texttt{IndexError} (judged: runtime error). \\
\bottomrule
\end{tabularx}
\caption{Agent prompts and example answers (target documents) in the programming domain.}
\label{tab:q1b}
\end{table*}

\begin{table*}[p]
\centering\scriptsize
\setlength{\tabcolsep}{5pt}
\renewcommand{\arraystretch}{1.08}
\begin{tabularx}{\textwidth}{@{}>{\raggedright\arraybackslash\itshape}p{0.105\textwidth}>{\raggedright\arraybackslash}X@{}}
\toprule
\multicolumn{2}{@{}l}{\textbf{CaseHOLD}} \\
Message & Read question.txt in this directory and answer it. The corpus is the pages/ directory. Write your final answer to answer.txt as a JSON array of record ids. \\
\cmidrule(l{0pt}r{0pt}){1-2}
Question & Answer the question below about the records in this collection. Each record is a citing text: a passage from a US court decision in which the court cites earlier cases; the first line of each record gives its record id. In every record, one citation is followed by a holding statement marked with double square brackets, for example \textquotedbl{}([[holding that the district court did not abuse its discretion in denying the motion]])\textquotedbl{}. Only the marked holding statement is in question. The cited cases themselves are not in the collection.\par
Question. In most records the marked holding statement is the correct one: the holding statement that corresponds to the citing text. In the other records it is an incorrect one: another holding statement, which does not correspond to this citing text. Find every record whose marked holding statement is incorrect.\par
Write your final answer to answer.txt as a JSON array of record ids, exactly as each record\textquotesingle{}s first line gives them (for example \textquotedbl{}h0a1b2c3d4\textquotedbl{}). \\
\cmidrule(l{0pt}r{0pt}){1-2}
Example answer & \emph{Record h578393828.} of the anchor. If the damages “arose out of the defective and unsafe condition” of the anchor, Jaenty’s claims are barred by the two-{}year statute of limitations. Jaenty argues that its injuries arose out of “negligent construction activities” rather than the defective and unsafe condition of the anchor and thus Jaenty’s claims should not be subject to the two-{}year statute of limitations. We address this issue in two ways: (1) whether the condition of the anchor was “defective and unsafe,” and (2) whether Jaenty’s alleged injuries “arose out of the condition” of the anchor. We have previously said that whether an injury arises out of a defective and unsafe condition “turns on the individual facts alleged in the complaint.” Gribel v. Andersen Corp., 489 N.W.2d 521, 523 (Minn.1992) ([[holding that a remedy for unsafe conditions need not await a tragic event]]). Here, the complaint alleged that the anchor \\
\cmidrule(l{0pt}r{0pt}){1-2}
Why it qualifies & \emph{Gribel} is cited for when a condition is defective and unsafe; the marked statement is another case's holding. CaseHOLD's correct holding is ``holding that to be defective and unsafe an improvement need not be a risk to human health.'' \\
\bottomrule
\end{tabularx}
\caption{Agent prompts and example answers (target documents) in the legal domain: CaseHOLD.}
\label{tab:q1b2}
\end{table*}

\begin{table*}
\centering\scriptsize
\setlength{\tabcolsep}{5pt}
\renewcommand{\arraystretch}{1.08}
\begin{tabularx}{\textwidth}{@{}>{\raggedright\arraybackslash\itshape}p{0.105\textwidth}>{\raggedright\arraybackslash}X@{}}
\toprule
\multicolumn{2}{@{}l}{\textbf{CLERC}} \\
Message & Read question.txt in this directory and answer it. The corpus is the pages/ directory. Write your final answer to answer.txt as one JSON object with the keys \textquotedbl{}1\textquotedbl{}, \textquotedbl{}2\textquotedbl{}, \textquotedbl{}3\textquotedbl{} and \textquotedbl{}4\textquotedbl{}, each a JSON array of record ids. \\
\cmidrule(l{0pt}r{0pt}){1-2}
Question & Answer four questions about the passages in this collection. Each passage is a block of text of 300 words taken from a case document, an opinion of a United States federal court; its first line gives its passage id. The citation in the middle of the block, its central citation, has been removed, together with every other reference to the case it cites: every citation to that case (its full citation, short forms such as \textquotedbl{}123 U.S. at 45\textquotedbl{}, and \textquotedbl{}Id.\textquotedbl{} references to it) is replaced by [CITATION], and every mention of that case\textquotesingle{}s name by [CASE]. Every other case cited in the passage is left as it is, so a case that a passage still cites or names is not the removed one.\par
Question 1. Find every passage whose central citation cites Terry v. Ohio, 392 U.S. 1 (1968).\par
Question 2. Find every passage whose central citation cites Illinois v. Gates, 462 U.S. 213 (1983).\par
Question 3. Find every passage whose central citation cites United States v. Leon, 468 U.S. 897 (1984).\par
Question 4. Find every passage whose central citation cites Schneckloth v. Bustamonte, 412 U.S. 218 (1973).\par
The collection also holds the case documents of these four cases, each in a document whose first line begins \textquotedbl{}Reference opinion\textquotedbl{}. They are not passages and belong to no answer.\par
Write your final answer to answer.txt as one JSON object with the keys \textquotedbl{}1\textquotedbl{}, \textquotedbl{}2\textquotedbl{}, \textquotedbl{}3\textquotedbl{} and \textquotedbl{}4\textquotedbl{}, each a JSON array of passage ids, exactly as each passage\textquotesingle{}s first line gives them (for example \textquotedbl{}c0a1b2c3d4\textquotedbl{}). \\
\cmidrule(l{0pt}r{0pt}){1-2}
Example answer & \emph{Question 3 (\emph{Leon}).} For ruling on a contested search made pursuant to a warrant, an alternative test is employed, as more fully discussed infra. First, the court determines ``whether the good-faith exception to the exclusionary rule applies''; if it does not, it must ascertain ``whether the warrant was supported by probable cause''. United States v. Laury, 985 F.2d 1293, 1311 (5th Cir.1993). The good-faith exception is applied unless: the issuing-judge was ``misled by information in an affidavit that the affiant knew was false or would have known was false except for his reckless disregard of the truth''; the issuing-judge ``wholly abandoned his judicial role'' in such a manner that ``no reasonably well trained officer should rely on the warrant''; the warrant was ``based on an affidavit so lacking in indicia of probable cause as to render official belief in its existence entirely unreasonable''; or the warrant was facially invalid. [CITATION] (internal quotation marks omitted). After hearing testimony by Agent Bailey and Gibbs, the magistrate judge called for supplemental briefing on ``the effect of not knowing what the evidence was that was offered in support of the warrant, [the second affidavit,] when all we have is the warrant itself'. In his supplemental brief, Gibbs reiterated his pre-hearing assertions: the second affidavit did not provide probable cause for a warrant; and his statements at the time of his arrest were involuntary. Concerning the alternative test for ruling on a contested search made pursuant to a warrant, Gibbs contended: because, at the suppression hearing, he had shown by a preponderance of the evidence that Agent Bailey had not acted in good faith in relying on the warrant, the burden had shifted to the Government to prove the sufficiency of the lost second \\
\cmidrule(l{0pt}r{0pt}){1-2}
Why it qualifies & These are \emph{Leon}'s exceptions to the good-faith rule; \emph{Leon} is never named. \\
\midrule
\multicolumn{2}{@{}l}{\textbf{CUAD}} \\
Message & Read question.txt in this directory and answer it. The corpus is the pages/ directory. Write your final answer to answer.txt as one JSON object with the keys \textquotedbl{}1\textquotedbl{}, \textquotedbl{}2\textquotedbl{}, \textquotedbl{}3\textquotedbl{} and \textquotedbl{}4\textquotedbl{}, each a JSON array of record ids. \\
\cmidrule(l{0pt}r{0pt}){1-2}
Question & Answer four questions about the clauses in this collection. Each document is one clause from a commercial contract filed with the US Securities and Exchange Commission; its first line gives its document id. Where a clause is made of text segments that do not appear immediately next to each other in the contract, the segments are joined by the symbol \textquotedbl{}\textless{}omitted\textgreater{}\textquotedbl{}.\par
Each question names a category of contract clause and gives its description. For each question, find every clause related to that category that should be reviewed by a lawyer.\par
Question 1. Category \textquotedbl{}Exclusivity\textquotedbl{}. Details: Is there an exclusive dealing commitment with the counterparty? This includes a commitment to procure all “requirements” from one party of certain technology, goods, or services or a prohibition on licensing or selling technology, goods or services to third parties, or a prohibition on collaborating or working with other parties), whether during the contract or after the contract ends (or both).\par
Question 2. Category \textquotedbl{}Post-{}Termination Services\textquotedbl{}. Details: Is a party subject to obligations after the termination or expiration of a contract, including any post-{}termination transition, payment, transfer of IP, wind-{}down, last-{}buy, or similar commitments?\par
Question 3. Category \textquotedbl{}Revenue/Profit Sharing\textquotedbl{}. Details: Is one party required to share revenue or profit with the counterparty for any technology, goods, or services?\par
Question 4. Category \textquotedbl{}Governing Law\textquotedbl{}. Details: Which state/country\textquotesingle{}s law governs the interpretation of the contract?\par
Write your final answer to answer.txt as one JSON object with the keys \textquotedbl{}1\textquotedbl{}, \textquotedbl{}2\textquotedbl{}, \textquotedbl{}3\textquotedbl{} and \textquotedbl{}4\textquotedbl{}, each a JSON array of document ids, exactly as each document\textquotesingle{}s first line gives them (for example \textquotedbl{}c0a1b2c3d4\textquotedbl{}). \\
\cmidrule(l{0pt}r{0pt}){1-2}
Example answer & \emph{Question 3 (Revenue/Profit Sharing).} Excite@Home and e-{}centives shall share equally all net revenue from           such offers, defined as gross revenue minus third-{}party serving           costs, which shall not exceed \$0.01 per email piece delivered. \\
\cmidrule(l{0pt}r{0pt}){1-2}
Why it qualifies & The two parties share the net revenue from the offers equally. \\
\bottomrule
\end{tabularx}
\caption{Agent prompts and example answers (target documents) in the legal domain: CLERC and CUAD.}
\label{tab:q1c1}
\end{table*}

\begin{table*}[p]
\centering\scriptsize
\setlength{\tabcolsep}{5pt}
\renewcommand{\arraystretch}{1.08}
\begin{tabularx}{\textwidth}{@{}>{\raggedright\arraybackslash\itshape}p{0.105\textwidth}>{\raggedright\arraybackslash}X@{}}
\toprule
\multicolumn{2}{@{}l}{\textbf{Evidence Inference}} \\
Message & Read question.txt in this directory and answer it. The corpus is the pages/ directory. Write your final answer to answer.txt as a JSON array of record ids. \\
\cmidrule(l{0pt}r{0pt}){1-2}
Question & Each record is one comparison from a clinical report of a randomized controlled trial: a COMPARISON part naming a medical intervention, a comparator and an outcome, followed by the ARTICLE part, the full text of the trial report. The first line of each record gives its record id.\par
For each record: with respect to the outcome, characterize the reported difference between patients receiving the intervention and those receiving the comparator, as significantly increased, significantly decreased or no significant difference. Find every record where it is significantly decreased. Note that if the outcome results in a benefit, such as more pain relief or increased weight loss then the intervention counts as an increase.\par
Write your final answer to answer.txt as a JSON array of record ids, exactly as each record\textquotesingle{}s first line gives them (for example \textquotedbl{}e0a1b2c3d4\textquotedbl{}). \\
\cmidrule(l{0pt}r{0pt}){1-2}
Example answer & \emph{Record ee74317f8a, complete (1,164 words).} COMPARISON\newline Intervention: Tissue plasminogen activator + deoxyribonuclease\newline Comparator: Placebo\newline Outcome: Hospital stay\newline ARTICLE\newline TITLE: New strategies to manage complicated pleural effusions\newline   ABSTRACT.EXPANDED ABSTRACT.CITATION:\newline Rahman NM, Maskell NA, West A, Teoh R, Arnold A, Mackinlay C, Peckham D, Davies CW, Ali N, Kinnear W, Bentley A, Kahan BC, Wrightson JM, Davies HE, Hooper CE, Lee YC, Hedley EL, Crosthwaite N, Choo L, Helm EJ, Gleeson FV, Nunn AJ, Davies RJ: N Engl J Med 2011, 365:518-{}26. PMID: 21830966, available on www.pubmed. gov\newline ABSTRACT.EXPANDED ABSTRACT.BACKGROUND:\newline More than 30\% of patients with pleural infection either die or require surgery. Drainage of infected fluid is the key to successful treatment, but intrapleural fibrinolytic therapy did not improve outcomes in an earlier, large, randomized trial therapy (Multicenter Intrapleural Sepsis Trial [MIST1]).\newline ABSTRACT.EXPANDED ABSTRACT.METHODS:\newline Objective: To evaluate the efficacy and safety of intrapleural DNase alone, alteplase alone, or the combination of both, to improve pleural drainage.  Design: Multicenter, double-{}blind, double-{}dummy, 2 × 2 factorial randomized trial.  Setting: Eleven centers in the United Kingdom (UK).  Subjects: Adult patients (mean age 59 years, 72\% men), who had clinical evidence of infection, and pleural fluid that had macroscopic purulence, a positive culture or Gram stain for bacteria, or a pH \textless{} 7.2.  Intervention: Patients were assigned to 1 of the 4 study interventions for 3 days: double placebo, intrapleural tissue plasminogen activator (t-{}PA) and DNase, t-{}PA and placebo, or DNase and placebo.  Outcomes: The primary outcome was the change in pleural opacity, measured as the percentage of the hemithorax occupied by effusion, on chest radiography on day 7 as compared with day 1. Secondary outcomes included referral for surgery, duration of hospital stay, and adverse events.\newline ABSTRACT.EXPANDED ABSTRACT.RESULTS:\newline The mean (± SD) change in pleural opacity was greater in the t-{}PA-{}DNase group than in the placebo group (-{}29.5 ± 23.3\% vs. -{}17.2 ± 19.6\%; difference, -{}7.9\%; 95\% confidence interval [CI], -{}13.4 to -{}2.4; P = 0.005). The change observed with t-{}PA alone and with DNase alone (-{}17.2 ± 24.3 and -{}14.7 ± 16.4\%, respectively) was not significantly different from that observed with placebo. The frequency of surgical referral at 3 months was lower in the t-{}PA-{}DNase group than in the placebo group (2 of 48 patients [4\%] vs. 8 of 51 patients [16\%]; odds ratio for surgical referral, 0.17; 95\% CI, 0.03 to 0.87; P = 0.03) but was greater in the DNase group (18 of 46 patients [39\%]) than in the placebo group (odds ratio, 3.56; 95\% CI, 1.30 to 9.75; P = 0.01). Combined t-{}PA-{}DNase therapy was associated with a reduction in the hospital stay, as compared with placebo (difference, -{}6.7 days; 95\% CI, -{}12.0 to -{}1.9; P = 0.006). Hospital stay with either agent alone was not significantly different from that with placebo. The frequency of adverse events did not differ significantly among the groups.\newline ABSTRACT.EXPANDED ABSTRACT.CONCLUSIONS:\newline Intrapleural t-{}PA-{}DNase therapy improved fluid drainage in patients with pleural infection and reduced the frequency of surgical referral and the duration of hospital stay. Treatment with DNase alone or t-{}PA alone was ineffective.\newline BODY.COMMENTARY:\newline More than 65,000 cases of pleural infection occur in the United States and United Kingdom each year [1], and the incidence is increasing in both children and adults [2]. Both community and hospital acquired bacterial pneumonia are complicated by pleural infections [3]. Despite the advances in therapeutic strategies, pleural infections are associated with high mortality rates of 10-{}20\% [4].  There are several classifications of pleural effusions based on evolution (exudative, fibrinoproliferative with evidence of loculation, and formation of fibrous peel) and fluid characteristics (clear, viscous, or pus). Drainage alone is usually adequate for an uncomplicated effusion. However, drainage is often insufficient for patients with pleural infection who have a complicated effusion with increased viscosity and loculation [5]. The standard management of pleural infections consists of antibiotics and draining the infected pleural fluid [6]. However, more than 30\% of patients fail standard therapy and require surgical referral. These patients incur a longer duration of hospital stay of 12-{}15 days and increased utilization of healthcare resources [4].  Observational data suggest that using fibrinolytic drugs in the intrapleural space can decrease the failure rate, reduce surgical referral, and improve chest tube drainage by cleaving intrapleural fibrinous septations [7,8]. This theory was tested by many randomized controlled trials using different fibrinolytics, but failed to improve clinically meaningful end points [9]. Most notably, the 454 patients in the Multicenter Intrapleural Sepsis Trial (MIST1) showed no benefit from intrapleural streptokinase. However, the absence of benefit was felt possibly due to the presence of free extracellular deoxyribonucleic acid (DNA) and other bacterial components in the pleural space, which increase fluid viscosity and hinder fibrinolysis [8]. Adding DNase to cleave the free DNA may reduce fluid viscosity and permit pleural clearance by fibrinolytic drugs [10].  To test this hypothesis, the second multicenter trial (MIST2) by Rahman and colleagues was conducted at 11 centers in the UK. In this randomized trial, the authors used a different intrapleural fibrinolytic agent (t-{}PA rather than streptokinase) and added DNase for cleavage of free DNA. The authors conducted a standard factorial study analysis and found a significant interaction between t-{}PA and DNase with regard to the primary outcome. They therefore compared each of the three intervention groups to placebo.  The t-{}PA-{}DNase combination was significantly superior to placebo and each drug alone in improving the changes in opacity on chest x-{}ray on day 7 vs. day 1 (primary end point), decreasing the number of surgical referrals at 3 and 12 months (4\% vs. 14\%), and decreasing the days of hospitalizations by 6.7 days. However, monotherapy with either t-{}PA or DNase showed no benefit when compared to placebo, and DNase alone resulted in an increase in surgical referrals and hospitalization days compared to placebo.  Strengths of this trial include: a well-{}defined study protocol, multicenter, randomized, double-{}blind, double-{}dummy design, and complete follow-{}up. There are a few limitations. First, although only 3 serious adverse events with pulmonary bleeding (intrapleural hemorrhage, hemoptysis) occurred, all three were in the t-{}PA-{}DNase group (n = 52). Future use of this drug combination should carefully monitor for pulmonary bleeding. Second, the primary end point was the absolute change in chest x-{}ray opacity. However, the trial was powered to detect a slightly different endpoint, the proportion of patients who had a 50\% relative reduction in opacity. Never-{}theless, the overall results were striking, with only one patient in the t-{}PA-{}DNase arm not achieving any opacification reduction, compared to several patients in the other arms. Additionally, there was reduction in the number of surgical referrals, a patient-{}centered endpoint.\newline BODY.RECOMMENDATION:\newline Rahman and colleagues should be commended for demonstrating that dual intrapleural therapy with t-{}PA and DNase in patients with pleural infection improved changes in opacity on chest x-{}ray and reduced need for surgical intervention and hospital stay. Although larger trials are needed to replicate these results, dual intrapleural therapy may be a useful option, especially for patients who are high-{}risk for surgery.\newline BODY.COMPETING INTERESTS:\newline The authors declare that they have no competing interests. \\
\cmidrule(l{0pt}r{0pt}){1-2}
Why it qualifies & A significantly shorter stay for exactly this comparison. \\
\bottomrule
\end{tabularx}
\caption{Agent prompts and example answers (target documents) in the medical domain: Evidence Inference.}
\label{tab:q2}\label{tab:q2b}
\end{table*}

\begin{table*}[p]
\centering\scriptsize
\setlength{\tabcolsep}{5pt}
\renewcommand{\arraystretch}{1.08}
\begin{tabularx}{\textwidth}{@{}>{\raggedright\arraybackslash\itshape}p{0.105\textwidth}>{\raggedright\arraybackslash}X@{}}
\toprule
\multicolumn{2}{@{}l}{\textbf{MedQA+}} \\
Message & Read question.txt in this directory and answer it. The corpus is the pages/ directory. Write your final answer to answer.txt as a JSON array of record ids. \\
\cmidrule(l{0pt}r{0pt}){1-2}
Question & Each record is one clinical text, a medical narrative about a patient; its first line gives its record id. Each clinical text is either correct or contains one error.\par
Find every clinical text that has a medical error related to treatment, management, cause, or diagnosis. Finding the error requires medical knowledge and reasoning.\par
Write your final answer to answer.txt as a JSON array of record ids, exactly as each record\textquotesingle{}s first line gives them (for example \textquotedbl{}m0a1b2c3d4\textquotedbl{}). \\
\cmidrule(l{0pt}r{0pt}){1-2}
Example answer & \emph{A MEDEC note.} A previously healthy 61-year-old man comes to the physician because of a 3-month history of intermittent fever, easy fatiguability, and a 4.4-kg (9.7-lb) weight loss. Physical examination shows conjunctival pallor. The spleen is palpated 5 cm below the left costal margin. Laboratory studies show a leukocyte count of 75,300/mm3 with increased basophils, a platelet count of 455,000/mm3, and a decreased leukocyte alkaline phosphatase score. Patient was diagnosed with acute promyelocytic leukemia after a peripheral blood smear showed increased numbers of promyelocytes, myelocytes, and metamyelocytes. \\
\cmidrule(l{0pt}r{0pt}){1-2}
Why it qualifies & The findings describe chronic myeloid leukemia and not acute promyelocytic leukemia (MEDEC's correction). \\
\midrule
\multicolumn{2}{@{}l}{\textbf{BioDEX}} \\
Message & Read question.txt in this directory and answer it. The corpus is the pages/ directory. Write your final answer to answer.txt as one JSON object with one key per question, exactly the keys question.txt lists, each a JSON array of record ids. \\
\cmidrule(l{0pt}r{0pt}){1-2}
Question & Answer five questions about the records in this collection. Each record is one published article (title, abstract and full text) that describes at least one adverse drug event; its first line gives its record id.\par
Each question names a patient reaction, as a MedDRA term (MedDRA is a standardized medical terminology; its terms are encoded in British English). For each question, read each article, extract information about the adverse drug event mentioned by the authors, and find every article in which that reaction was part of an adverse drug event.\par
Question 1. Reaction \textquotedbl{}Acute kidney injury\textquotedbl{}.\par
Question 2. Reaction \textquotedbl{}Neutropenia\textquotedbl{}.\par
Question 3. Reaction \textquotedbl{}Thrombocytopenia\textquotedbl{}.\par
Question 4. Reaction \textquotedbl{}Pyrexia\textquotedbl{}.\par
Question 5. Reaction \textquotedbl{}Respiratory failure\textquotedbl{}.\par
Write your final answer to answer.txt as one JSON object with the keys \textquotedbl{}1\textquotedbl{}, \textquotedbl{}2\textquotedbl{}, \textquotedbl{}3\textquotedbl{}, \textquotedbl{}4\textquotedbl{} and \textquotedbl{}5\textquotedbl{}, each a JSON array of record ids, exactly as each record\textquotesingle{}s first line gives them (for example \textquotedbl{}r0a1b2c3d4\textquotedbl{}). \\
\cmidrule(l{0pt}r{0pt}){1-2}
Example answer & \emph{Question 1 (Acute kidney injury); a 764-{}word article.} Vasodilation through levodopa for Parkinson\textquotesingle{}s disease may require high left ventricular assist device flow.\newline We report implantation of a left ventricular assist device (LVAD) in a patient with Parkinson\textquotesingle{}s disease. Postoperative fluid overload together with insufficient LVAD output in the setting of vasodilation through levodopa likely caused renal hypoperfusion and acute kidney injury. \ldots{} Two weeks after surgery, he was readmitted to the ICU with pulmonary edema and anuric KDIGO stage 3 acute kidney injury (AKI). \\
\cmidrule(l{0pt}r{0pt}){1-2}
Why it qualifies & The authors attribute the patient's acute kidney injury to vasodilation through levodopa. \\
\bottomrule
\end{tabularx}
\caption{Agent prompts and example answers (target documents) in the medical domain: MedQA+ and BioDEX.}
\label{tab:q2c1}
\end{table*}

%% file: content/appendix/agents.tex
\section{Evaluation Setup and Cost}
\label{app:agents}

\subsection{Evaluation Setup}
\label{app:resources}
We evaluate GPT-6.1-Sol and GPT-6-Astra with Codex, and Opus~5.5 with Claude Code, all set to \texttt{xhigh} reasoning effort.

\label{app:prompts}
Each run starts in a fresh workspace containing \texttt{question.txt} and a \texttt{pages/} directory with one file per document. 
Each file name is a string of 8 random hexadecimal digits.
Tables~\ref{tab:prompts} and~\ref{tab:answer-formats} specify the prompt template and task-specific answer formats.

\begin{table*}[!t]
  \centering\small
  \setlength{\tabcolsep}{7pt}
  \renewcommand{\arraystretch}{1.15}
  \begin{tabularx}{\textwidth}{@{}>{\raggedright\arraybackslash}p{0.20\textwidth}>{\raggedright\arraybackslash}X@{}}
    \toprule
    \textbf{Agents} & \textbf{Prompt Text} \\
    \midrule
     Claude Code, Codex & \texttt{Read question.txt in this directory and answer it. The corpus is the pages/ directory. Write your final answer to answer.txt} {[FORMAT]}\texttt{.} \\\midrule
    \addlinespace[5pt]
    Exhaustive Search Baseline & \texttt{Read the question below and answer it. The corpus is the documents below. Write your final answer} {[FORMAT]}\texttt{.}, then a blank line, the text of \texttt{question.txt}, a blank line, and the call's documents separated by blank lines \\
    \bottomrule
  \end{tabularx}
  \caption{The \dataset{} search task prompt template. Replace [FORMAT] with the task-specific string from Table~\ref{tab:answer-formats}. The bottom prompt is used for the exhaustive search agent baseline in \S\ref{app:luna}}
  \label{tab:prompts}

  \vspace{12pt}
  \begin{tabularx}{\textwidth}{@{}>{\raggedright\arraybackslash}p{0.20\textwidth}>{\raggedright\arraybackslash}X@{}}
    \toprule
    \textbf{Task} & \textbf{Format String} \\
    \midrule
    NLI, Codeforces + CodeNet, CaseHOLD, Evidence Inference, MedQA+ & \texttt{as a JSON array of record ids} \\\midrule
    \addlinespace[5pt]
    HoVer, FEVEROUS & \texttt{as a JSON array of claim labels} \\\midrule
    \addlinespace[5pt]
    CLERC, CUAD & \texttt{as one JSON object with the keys "1", "2", "3" and "4", each a JSON array of record ids} \\\midrule
    \addlinespace[5pt]
    BioDEX & \texttt{as one JSON object with one key per question, exactly the keys question.txt lists, each a JSON array of record ids} \\
    \bottomrule
  \end{tabularx}
  \caption{The answer format strings inserted into the individual tasks' prompt templates in Table~\ref{tab:prompts}.}
  \label{tab:answer-formats}
\end{table*}

\subsection{Corpus Samples}
\label{app:corpora}
For each task, the full corpus is the document pool listed in Table~\ref{tab:benchmarks}.
The ten 50\% corpora are independent random 50\% samples drawn with distinct seeds, so they overlap and vary slightly in their number of answers.
The remaining sizes form a nested chain: we shuffle the full corpus once with a fixed seed and take the first 10\%, 25\%, 75\%, and 100\% of documents, ensuring each smaller corpus is contained within the larger ones.
Repeated runs at a given size use the same corpus.
For the 50\% point of the corpus-size experiment (Figure~\ref{fig:sizes-main} and Table~\ref{tab:sizes}), each agent runs three times on the first of the ten 50\% corpora.
Each corpus retains the original document texts, labels, and task question, with its answer key restricted to the included documents.

\subsection{Exhaustive Search Agent Baseline}
\label{app:luna}
We evaluate an exhaustive search agent as a brute-force baseline, iterating over every document in the corpus. We use GPT-6-Luna with high reasoning effort as our reader LLM. It reads the entire corpus in batches of 20 documents, capped at a total length of 60K tokens (OpenAI tokenizer). GPT-6-Luna is prompted using the prompt in Table~\ref{tab:prompts}, followed by the individual task prompt described in \S\ref{app:tasks}.

Due to budget constraints, this baseline is evaluated over a single 50\% sample of \dataset{}, consisting of 10 search tasks in total. Results are displayed in Figure~\ref{fig:sizes-main}, while Table~\ref{tab:luna-cost} reports precision, recall, and F1 scores as well as the total cost in terms of: LLM calls, new input + output tokens, and monetary cost.

To compute the cost estimates, we price our submitted prompt tokens based on GPT-6-Luna's input pricing at \$0.10 per 1M tokens (as of November 2026). All remaining tokens, including Codex's own instructions, reasoning tokens, and output, at its using its output pricing (\$0.50 per 1M tokens). Since part of Codex's instructions is likely served from cache, this cost estimate serves as an upper bound.

\subsection{Evaluation Cost}
\label{app:cost}
We describe the exact details for the cost estimates and token usage by each of the evaluated agents in \S\ref{sec:experiments:cost} and Table~\ref{tab:tokens}.

\paragraph{Pricing.}
The cost (in USD) of each agent run on a single \dataset{} task is calculated by summing the costs of all LLM calls made during the run, including those made by sub-agents. Pricing is based on the standard API rates of OpenAI and Anthropic as of November 2026, accounting for prompt caching, input and output tokens, and reasoning tokens (which are billed as output tokens).

\begin{figure*}
  \centering
  \includegraphics[width=\textwidth]{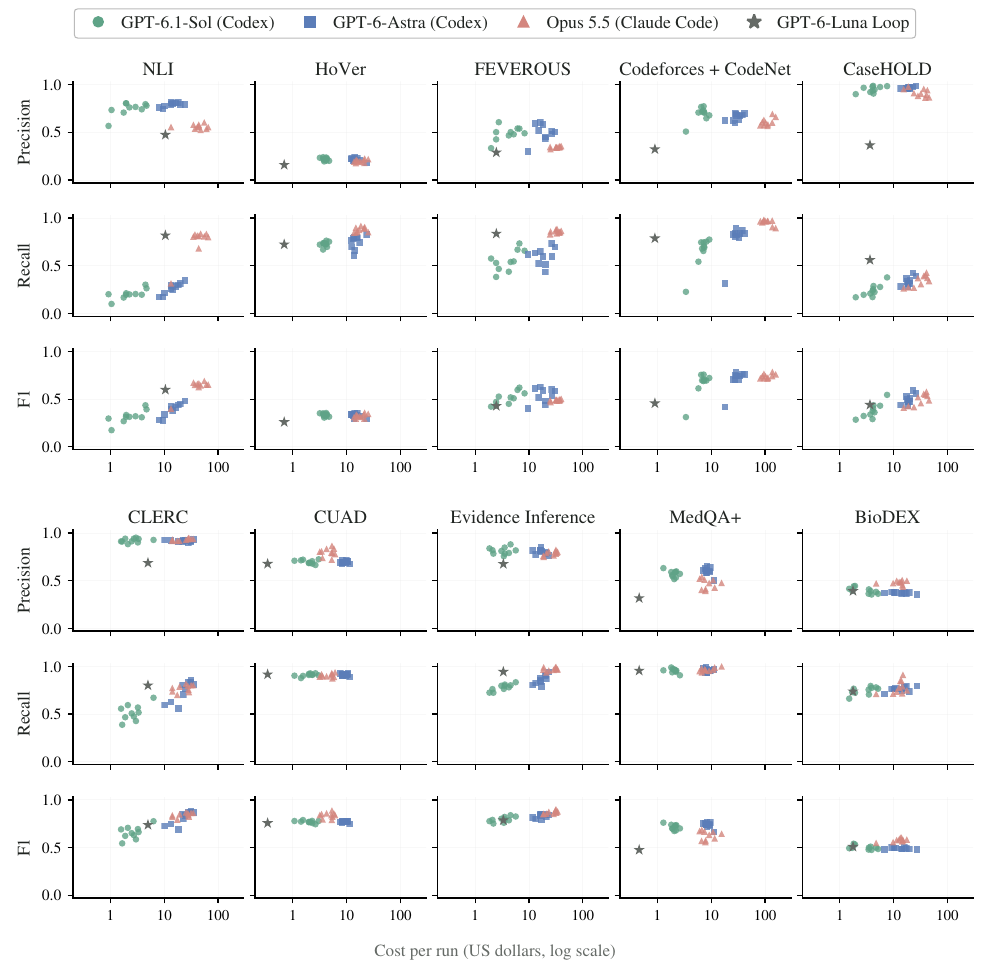}
  \caption{Agent performance as a function of cost for each \dataset{} task. Each marker is one run, with 10 runs per agent for each task. The star denotes the exhaustive agent baseline (\S\ref{app:luna}) and its cost from Table~\ref{tab:luna-cost}.}
  \label{fig:prf-cost}
\end{figure*}

To illustrate the cost-effectiveness of each agent, Figure~\ref{fig:prf-cost} plots per-task precision, recall, and F1 scores across a 50\% sample of \dataset{} as a function of the cost per run in USD.

\paragraph{Tokens.}
We describe the different token categories used by the agent during its execution. \emph{Cached input} refers to the portion of a prompt served from the provider's cache, typically consisting of content from previous conversation turns, whereas \emph{new input} refers to input tokens that are not cached.

For OpenAI models, we calculate new input by subtracting the number of cached input tokens from the total number of input tokens. For Anthropic models, we calculate new input as the sum of ``uncached input'' tokens and ``cache writes'', since Anthropic reports uncached input, cache writes, and cache reads separately. 
Finally, \emph{(new input + output) / corpus tokens} divides the sum of the new input and output tokens, (including reasoning tokens but excluding cached input) by the corpus size. It measures the number of tokens consumed by the agent relative to the number of tokens required to read the entire corpus.
Token numbers are computed using OpenAI's o200k tokenizer.

\newpage
\subsection{Error Analysis}\label{app:experiments:error_analysis}
We wish to assess how accurate the examples in \dataset{} are in assessing agents' precision. To this end we conduct an error analysis of false positive results of all three agents: Opus~5.5, GPT-6.1-Sol and GPT-6-Astra. We take 100 random false positives per task (1,000 total) where a third belong to each of the three agents that we evaluated (340 for GPT-6.1-Sol). The results in Table~\ref{tab:label-audit} categorize each false positive example either as a dataset error, where the agent returned a target document, or as an actual error by the agent. We consider an example to be a dataset error by using an GPT-6.2-Sol as a judge the prompt in Table~\ref{tab:judge-prompt}.

Overall, 58.5\% of the false positive cases are judged to be labeling errors of the original dataset. We present two examples from HoVer and BioDEX in Table~\ref{tab:dataset-errors}. 
However, agents errors still constitute 41.5\% of the false positive examples, demonstrating that their precision remains far from perfect. Furthermore, since all agents are evaluated on the same set of examples, these issues would not affect the overall \emph{ranking} between the agents precision.

\begin{table}[t]
  \centering\footnotesize\setlength{\tabcolsep}{4pt}
  \input{content/tables/tab_label_audit.tex}
  \caption{The average False Positive (FP) averaged across all agents, together with an error analysis. For 100 random false positive per task, we examine whether the agents' were truly incorrect or whether there was a labeling error in the original dataset. Error breakdown per agent is provided in Table~\ref{tab:label-audit-agents}.}
  \label{tab:label-audit}
\end{table}

%% file: content/tables/tab_label_audit.tex
\begin{tabular}{@{}lrrr@{}}
\toprule
Task & \shortstack[l]{False Pos.\\Prevalence} & \shortstack[l]{Dataset Err.\\(\% of FP)} & \shortstack[l]{Agent Err.\\(\% of FP)} \\
\midrule
NLI & 1.0\% & 61\% & 39\% \\
HoVer & 12.0\% & 95\% & 5\% \\
FEVEROUS & 4.0\% & 67\% & 33\% \\
Codeforces+CodeNet & 2.3\% & 80\% & 20\% \\
CaseHOLD & 0.1\% & 63\% & 37\% \\
CLERC & 0.3\% & 0\% & 100\% \\
CUAD & 2.4\% & 59\% & 41\% \\
Evidence Inference & 1.1\% & 55\% & 45\% \\
MedQA+ & 3.6\% & 48\% & 52\% \\
BioDEX & 6.6\% & 57\% & 43\% \\
\midrule
\textbf{Average} & 3.4\% & 58.5\% & 41.5\% \\
\bottomrule
\end{tabular}

%% file: content/appendix/results.tex
\begin{table*}[!t]
  \centering
  \footnotesize\setlength{\tabcolsep}{3.5pt}
  \input{content/tables/tab_results.tex}
  \caption{Agents results on the 100-document results and on the ten randomly sampled corpora (lowest and highest scores are in parentheses). \dataset{} results, in the bottom row, present the average results over all tasks.}
  \label{tab:results}
\end{table*}

\begin{table*}[!t]
  \centering
  \scriptsize\setlength{\tabcolsep}{2.5pt}
  \input{content/tables/tab_sizes_25_100.tex}
  \caption{The agents' performance on varying sizes of the document corpus. For each sample (10\%, 25\%, 50\%, 75\%, 100\%) the tables results are averaged over 3 separate agent runs per sample. \dataset{} results, in the bottom row, present the average results over all ten exhaustive search tasks.}
  \label{tab:sizes}
\end{table*}

\begin{table*}
  \centering\footnotesize\setlength{\tabcolsep}{5pt}
  \input{content/tables/tab_luna_cost.tex}
  \caption{Exhaustive agent baseline results over a 50\% corpus sample of \dataset{}. Token numbers computed using the OpenAI tokenizer. \emph{Input + output} is the amount of input, reasoning and output tokens used by the LLM (exc. cached tokens). The cost estimate is derived from GPT-6-Luna pricing, described in \S\ref{app:luna}.}
  \label{tab:luna-cost}\label{tab:luna}
\end{table*}

\begin{table*}[h]
  \centering\footnotesize\setlength{\tabcolsep}{5pt}
  \input{content/tables/tab_tokens.tex}
  \caption{Agents' average cost and token usage across \dataset{} tasks. Each task row is averaged across 10 runs, each over a distinct 50\% sample of the task corpus. The \dataset{} (total) row sums these per-task averages across all 10 tasks. \emph{(Input + out.) / corpus} represents new input, reasoning, and output tokens divided by the corpus size.}
  \label{tab:tokens}
  
\end{table*}

\begin{table*}[ht]
\centering\scriptsize
\setlength{\tabcolsep}{5pt}
\renewcommand{\arraystretch}{1.08}
\begin{tabularx}{\textwidth}{@{}>{\raggedright\arraybackslash\itshape}p{0.105\textwidth}>{\raggedright\arraybackslash}X@{}}
\toprule
\multicolumn{2}{@{}l}{\textbf{HoVer}} \\
Question & Find every claim that is not supported by the articles in its file. \\
\cmidrule(l{0pt}r{0pt}){1-2}
Document & \emph{Claim 132314.} Pachystegia and Passiflora, are not in the family of Passifloraceae.\newline Articles:\newline [Pachystegia]\newline Pachystegia is a genus of shrubs in the daisy family described as a genus in 1925.\newline [Passiflora]\newline Passiflora, known also as the passion flowers or passion vines, is a genus of about 500 species of flowering plants, the type genus of the family Passifloraceae. \ldots{} \\
\cmidrule(l{0pt}r{0pt}){1-2}
Dataset label & Supported (not an answer). \\
\cmidrule(l{0pt}r{0pt}){1-2}
Judge & Label wrong: ``Passiflora belongs to Passifloraceae, so the claim excluding both genera from that family fails.'' \\
\midrule
\multicolumn{2}{@{}l}{\textbf{BioDEX}} \\
Question & Reaction ``Thrombocytopenia'': find every article in which that reaction was part of an adverse drug event. \\
\cmidrule(l{0pt}r{0pt}){1-2}
Document & \emph{Record r9d0c14a53.} A Retrospective Analysis of R-MPV Plus Response-adapted Whole-brain Radiotherapy for Elderly Patients with Primary Central Nervous System Lymphoma. \ldots{} Hematological adverse events of grade $>$3 during induction were neutropenia (Patients 2-5, 7-10) (n=8), anemia (Patients 2, 5, and 10) (n=3), and thrombocytopenia (Patient 10) (n=1). Adverse events during high-dose AraC treatment were neutropenia (Patients 1-4, 6, 7) (n=6), anemia (Patients 4 and 6) (n=2), and thrombocytopenia (Patients 1, 3, 4 \ldots{} \\
\bottomrule
\end{tabularx}
\caption{Examples of dataset errors. False positives that the LLM judge attributes to a dataset error, with excerpts from the documents and the judge's stated reason.}
\label{tab:dataset-errors}
\end{table*}

\begin{table*}[t]
\centering\scriptsize
\setlength{\tabcolsep}{5pt}
\renewcommand{\arraystretch}{1.08}
\begin{tabularx}{\textwidth}{@{}>{\raggedright\arraybackslash}X@{}}
\toprule
\textbf{Prompt} \\
\midrule
You are reviewing the answer key of a dataset. question.txt is the task question exactly as the solver agents saw it. items.json lists the items: for each, the question number that applies, the document file in pages/, and the dataset label for that document (ANSWER = the dataset says this document qualifies; NOT AN ANSWER = the dataset says it does not). Some items also have original\_file: the same passage before its central citation was masked. For EVERY item, read the document yourself and decide whether the dataset label is right under the question's own definition, the way a careful expert following the source benchmark's definition would. Read each item; do not decide by keyword rules. For programming submissions you may run them on small valid inputs (inside this directory, no network). Verdicts: LABEL\_RIGHT, LABEL\_WRONG, or AMBIGUOUS (only when the document genuinely does not settle it). Write review.json in this directory as \{\textquotedbl{}items\textquotedbl{}: [\{\textquotedbl{}n\textquotedbl{}: 1, \textquotedbl{}verdict\textquotedbl{}: \textquotedbl{}LABEL\_WRONG\textquotedbl{}, \textquotedbl{}quote\textquotedbl{}: \textquotedbl{}short verbatim quote from the document that decides it\textquotedbl{}, \textquotedbl{}reason\textquotedbl{}: \textquotedbl{}one sentence\textquotedbl{}\}, ...]\} with one entry per item, every item covered. Rewrite review.json every 25 or so items so progress is saved. \\
\bottomrule
\end{tabularx}
  \caption{Prompt for the GPT-6.2-Sol judge (high reasoning).}
  \label{tab:judge-prompt}
\end{table*}

\begin{table*}[t]
  \centering\footnotesize\setlength{\tabcolsep}{4pt}
  \input{content/tables/tab_label_audit_agents.tex}
  \caption{Breakdown of Table~\ref{tab:label-audit}'s dataset errors by dataset and agent.}
  \label{tab:label-audit-agents}
\end{table*}

%% file: content/tables/tab_results.tex
\begin{tabular}{@{}llcccccc@{}}
\toprule
& & \multicolumn{3}{c}{100-document results} & \multicolumn{3}{c}{Ten sampled corpora (avg)} \\
\cmidrule(lr){3-5}\cmidrule(l){6-8}
Task & Agent & P & R & F1 & P & R & F1 \\
\midrule
NLI & GPT-6.1-Sol & 93.3 & 84.0 & 88.4 & 74.5\,{\scriptsize(56.6--80.3)} & 20.3\,{\scriptsize(9.9--30.1)} & 31.6\,{\scriptsize(17.5--43.6)} \\
 & GPT-6-Astra & 91.3 & 84.0 & 87.5 & 78.8\,{\scriptsize(74.3--81.2)} & 26.0\,{\scriptsize(17.1--34.3)} & 38.8\,{\scriptsize(27.9--47.9)} \\
 & Opus 5.5 & 89.8 & 88.0 & 88.9 & 55.8\,{\scriptsize(52.4--60.3)} & 75.4\,{\scriptsize(31.4--83.6)} & 63.2\,{\scriptsize(40.1--69.1)} \\
\midrule
HoVer & GPT-6.1-Sol & 84.9 & 90.0 & 87.4 & 21.8\,{\scriptsize(19.4--23.8)} & 72.6\,{\scriptsize(66.9--76.2)} & 33.4\,{\scriptsize(30.7--35.5)} \\
 & GPT-6-Astra & 84.6 & 88.0 & 86.3 & 21.3\,{\scriptsize(17.8--23.9)} & 74.7\,{\scriptsize(60.9--82.6)} & 33.0\,{\scriptsize(29.3--35.4)} \\
 & Opus 5.5 & 91.5 & 86.0 & 88.7 & 19.9\,{\scriptsize(17.7--22.4)} & 87.9\,{\scriptsize(84.8--91.7)} & 32.5\,{\scriptsize(29.6--35.8)} \\
\bottomrule
\end{tabular}

%% file: content/tables/tab_sizes_25_100.tex
\begin{tabular}{@{}ll ccc ccc ccc ccc ccc@{}}
\toprule
 & & \multicolumn{3}{c}{10\%} & \multicolumn{3}{c}{25\%} & \multicolumn{3}{c}{50\%} & \multicolumn{3}{c}{75\%} & \multicolumn{3}{c}{100\%} \\
\cmidrule(lr){3-5}\cmidrule(lr){6-8}\cmidrule(lr){9-11}\cmidrule(lr){12-14}\cmidrule(l){15-17}
Task & Agent & P & R & F1 & P & R & F1 & P & R & F1 & P & R & F1 & P & R & F1 \\
\midrule
NLI & GPT-6.1-Sol & 70.1 & 46.9 & 56.0 & 77.1 & 31.1 & 44.3 & 79.5 & 21.5 & 33.7 & 57.6 & 15.6 & 23.9 & 60.7 & 16.2 & 25.5 \\
 & GPT-6-Astra & 64.0 & 65.9 & 64.5 & 75.6 & 40.4 & 52.4 & 72.3 & 23.2 & 35.0 & 55.6 & 18.6 & 27.6 & 61.5 & 17.8 & 27.4 \\
 & Opus 5.5 & 56.4 & 83.1 & 67.1 & 56.8 & 80.3 & 66.4 & 55.7 & 80.8 & 65.9 & 56.5 & 81.4 & 66.7 & 54.4 & 79.9 & 64.7 \\
\midrule
HoVer & GPT-6.1-Sol & 22.0 & 89.2 & 35.3 & 22.3 & 82.7 & 35.1 & 22.8 & 68.1 & 34.1 & 26.8 & 49.8 & 34.7 & 26.4 & 56.9 & 35.9 \\
 & GPT-6-Astra & 21.7 & 91.4 & 35.1 & 21.0 & 82.7 & 33.5 & 21.2 & 79.7 & 33.5 & 22.1 & 71.5 & 33.8 & 25.5 & 69.0 & 37.2 \\
 & Opus 5.5 & 27.1 & 90.3 & 41.6 & 25.5 & 86.7 & 39.4 & 20.7 & 86.1 & 33.3 & 22.4 & 83.8 & 35.4 & 21.0 & 85.3 & 33.6 \\
\bottomrule
\end{tabular}

%% file: content/tables/tab_luna_cost.tex
\begin{tabular}{@{}lcccrrrrrr@{}}
\toprule
Task & P & R & F1 & Documents & \shortstack[r]{Model\\calls} & \shortstack[r]{Corpus\\tokens} & \shortstack[r]{New input\\+ output tok.} & \shortstack[r]{(Input + output)\\/ corpus tok.} & Cost est. \\
\midrule
NLI & 47.3 & 81.8 & 59.9 & 145,478 & 7,283 & 7.7M & 28.5M & 3.69$\times$ & \$10.50 \\
HoVer & 15.8 & 72.3 & 26.0 & 3,624 & 182 & 3.5M & 4.2M & 1.21$\times$ & \$0.69 \\
FEVEROUS & 28.8 & 83.6 & 42.9 & 25,252 & 1,263 & 3.6M & 8.2M & 2.28$\times$ & \$2.48 \\
Codeforces + CodeNet & 32.2 & 78.7 & 45.7 & 3,054 & 153 & 2.5M & 3.9M & 1.53$\times$ & \$0.89 \\
CaseHOLD & 36.4 & 56.0 & 44.1 & 25,290 & 1,265 & 6.4M & 12.6M & 1.98$\times$ & \$3.63 \\
CLERC & 68.6 & 80.1 & 73.8 & 30,068 & 1,525 & 12.6M & 20.9M & 1.66$\times$ & \$4.98 \\
CUAD & 67.7 & 91.7 & 75.7 & 4,153 & 208 & 0.3M & 1.0M & 3.01$\times$ & \$0.34 \\
Evidence Inference & 67.5 & 94.4 & 78.7 & 4,081 & 466 & 26.2M & 27.7M & 1.06$\times$ & \$3.34 \\
MedQA+ & 31.7 & 95.6 & 47.6 & 2,338 & 117 & 0.3M & 1.1M & 4.10$\times$ & \$0.46 \\
BioDEX & 39.2 & 73.8 & 50.9 & 3,554 & 225 & 13.0M & 14.3M & 1.10$\times$ & \$1.76 \\
\midrule
\textbf{\dataset{} (total)} & 43.5 & 80.8 & 54.5 & 246,892 & 12,687 & 76.1M & 122.5M & 1.61$\times$ & \$29.07 \\
\bottomrule
\end{tabular}

%% file: content/tables/tab_tokens.tex
\begin{tabular}{@{}lcccccccc@{}}
\toprule
Task & Cost & \shortstack[c]{Runtime \\(minutes)} & \shortstack[c]{Model\\calls} & \shortstack[c]{New input \\tokens} & \shortstack[c]{Cached \\tokens} & \shortstack[c]{Output \\tokens} & \shortstack[c]{Corpus\\tokens} & \shortstack[c]{(Input + out.) / \\corpus tok.} \\
\midrule
\multicolumn{9}{@{}l}{\textit{GPT-6.1-Sol (Codex)}} \\
NLI & \$2.58 & 28 & 63 & 0.27M & 9.27M & 0.06M & 7.76M & 0.04$\times$ \\
HoVer & \$3.97 & 41 & 74 & 0.38M & 13.9M & 0.06M & 3.51M & 0.13$\times$ \\
FEVEROUS & \$4.47 & 39 & 91 & 0.48M & 16.5M & 0.06M & 3.63M & 0.15$\times$ \\
Codeforces + CodeNet & \$6.87 & 45 & 115 & 0.84M & 22.9M & 0.07M & 2.54M & 0.36$\times$ \\
CaseHOLD & \$4.25 & 36 & 88 & 0.56M & 14.7M & 0.05M & 6.36M & 0.10$\times$ \\
CLERC & \$2.83 & 41 & 86 & 0.31M & 11.3M & 0.06M & 12.7M & 0.03$\times$ \\
\bottomrule
\end{tabular}

%% file: content/tables/tab_label_audit_agents.tex
\begin{tabular}{@{}lccccccccc@{}}
\toprule
& \multicolumn{3}{c}{GPT-6.1-Sol} & \multicolumn{3}{c}{GPT-6-Astra} & \multicolumn{3}{c}{Opus~5.5} \\
\cmidrule(lr){2-4} \cmidrule(lr){5-7} \cmidrule(lr){8-10}
Task & \shortstack[c]{False\\positives} & \shortstack[c]{Dataset\\errors} & \shortstack[c]{Dataset\\errors (\%)} & \shortstack[c]{False\\positives} & \shortstack[c]{Dataset\\errors} & \shortstack[c]{Dataset\\errors (\%)} & \shortstack[c]{False\\positives} & \shortstack[c]{Dataset\\errors} & \shortstack[c]{Dataset\\errors (\%)} \\
\midrule
NLI & 34 & 21 & 62\%  & 33 & 28 & 85\%  & 33 & 12 & 36\%  \\
HoVer & 34 & 33 & 97\%  & 33 & 32 & 97\%  & 33 & 30 & 91\%  \\
FEVEROUS & 34 & 26 & 76\%  & 33 & 21 & 64\%  & 33 & 20 & 61\%  \\
Codeforces + CodeNet & 34 & 27 & 79\%  & 33 & 26 & 79\%  & 33 & 27 & 82\%  \\
CaseHOLD & 34 & 23 & 68\%  & 33 & 24 & 73\%  & 33 & 16 & 48\%  \\
CLERC & 34 & 0 & 0\%  & 33 & 0 & 0\%  & 33  & 0 & 0\%  \\
CUAD & 34 & 19 & 56\%  & 33 & 20 & 61\%  & 33 & 20 & 61\%  \\
Evidence Inference & 34 & 20 & 59\%  & 33 & 21 & 64\%  & 33 & 14 & 42\%  \\
MedQA+ & 34 & 18 & 53\%  & 33 & 19 & 58\%  & 33 & 11 & 33\%  \\
BioDEX & 34 & 20 & 59\%  & 33 & 17 & 52\%  & 33 & 20 & 61\%  \\
\midrule
All tasks & 340 & 207 & 61\% & 330 & 208 & 63\% & 330 & 170 & 52\% \\
\bottomrule
\end{tabular}

%% file: content/appendix/coverage.tex
\section{Coverage Measurement}
\label{app:coverage}\label{app:coverage-definition}

To determine whether an agent read a document during a run, we measure document \emph{exposure} in the agent's execution trace. We collect all tool outputs produced during a run, including shell command outputs and printed file contents. For Claude Code, we also include the prompts sent to sub-agents and the outputs returned by their tools. We exclude the agent's own messages, commands, and reasoning traces, as these are generated by the model.

\paragraph{Exposure.}
Each document is formatted as newline-separated lines corresponding to individual fields or paragraphs. 
To extract identifying phrases from document~$i$, we lowercase all text, split each line into words, and omit document identifiers (so that printing an ID alone does not count as showing the document). 
Within each line, we collect every contiguous five-word n-gram—or the entire line if it is two to four words long—that appears in no other document in the corpus. 
Let $G_i$ denote this set of unique n-grams for document~$i$.

We process the run's tool outputs similarly by lowercasing and splitting them into words, but instead of splitting on newlines, we treat each tool output as a single continuous stream of words (allowing n-grams to span line breaks within an output, though never across separate tool outputs). 
Let $T$ be the set of all contiguous two- to five-word n-grams appearing in any tool output. Document~$i$'s exposure is then $c_i = \frac{|G_i \cap T|}{|G_i|}$.
Handling line breaks differently in documents and outputs makes detection robust to reformatting.
Restricting document n-grams to individual lines ensures they still match if an agent reorders fields or merges multiple lines into one.

\paragraph{Categories.}
Based on $c_i$, we classify each document as \textit{never shown} ($c_i = 0$), \textit{partly shown} ($0 < c_i < 0.5$), or \textit{mostly shown} ($c_i \geq 0.5$).

\paragraph{Main agent versus sub-agent exposure.}
For Claude Code, text shown to a Haiku~5.5 sub-agent counts as shown even if Opus~5.5 never sees it directly.
Table~\ref{tab:whom} separates these two by computing exposure independently from the main agent's tool outputs (which include sub-agent replies) and from the sub-agents' prompts and tool outputs.
Across the 100 Opus~5.5 runs, 79\% of documents are \textit{mostly shown} to Haiku~5.5 sub-agents compared with only 1\% to Opus~5.5 itself, and 78\% of its correct answers were shown to Haiku~5.5 but never to Opus~5.5.

\begin{table}[t]
  \centering\footnotesize\setlength{\tabcolsep}{3pt}
  \begin{tabular}{@{}lrrc@{}}
    \toprule
    \multirow{2}{*}{Task}
    & \multicolumn{2}{c}{\textit{Mostly shown} (\%)}
    & \multirow{2}{*}{\begin{tabular}{@{}c@{}}
        Correct answers\\
        \textit{never shown}\\
        to Opus~5.5 (\%)
      \end{tabular}} \\
    \cmidrule(lr){2-3}
    & \multicolumn{1}{c}{Opus~5.5}
    & \multicolumn{1}{c}{Haiku~5.5} & \\[7pt]
    \midrule
    NLI & 0.3 & 92 & 99 \\
    HoVer & 0.4 & 100 & 85 \\
    FEVEROUS & 0.4 & 100 & 99 \\
    Codeforces + CodeNet & 15.5 & 81 & 72 \\
    CaseHOLD & 1.3 & 81 & 36 \\
    CLERC & 0.1 & 6 & 65 \\
    CUAD & 21.6 & 76 & 26 \\
    Evidence Inference & 1.3 & 51 & 83 \\
    MedQA+ & 23.5 & 100 & 51 \\
    BioDEX & $<$0.1 & 2 & 95 \\
    \midrule
    All & 1.1 & 79 & 78 \\
    \bottomrule
  \end{tabular}
  \caption{
  Percentage of documents \textit{mostly shown} to Opus~5.5 and its Haiku~5.5 sub-agents across 100 runs on 50\% corpora, and the percentage of correct answers \textit{never shown} to Opus~5.5 but \textit{partly} or \textit{mostly shown} to Haiku~5.5. \emph{All} pools counts across all ten tasks.}
  \label{tab:whom}
\end{table}

\begin{figure*}[!t]
  \centering
  \includegraphics[width=\textwidth]{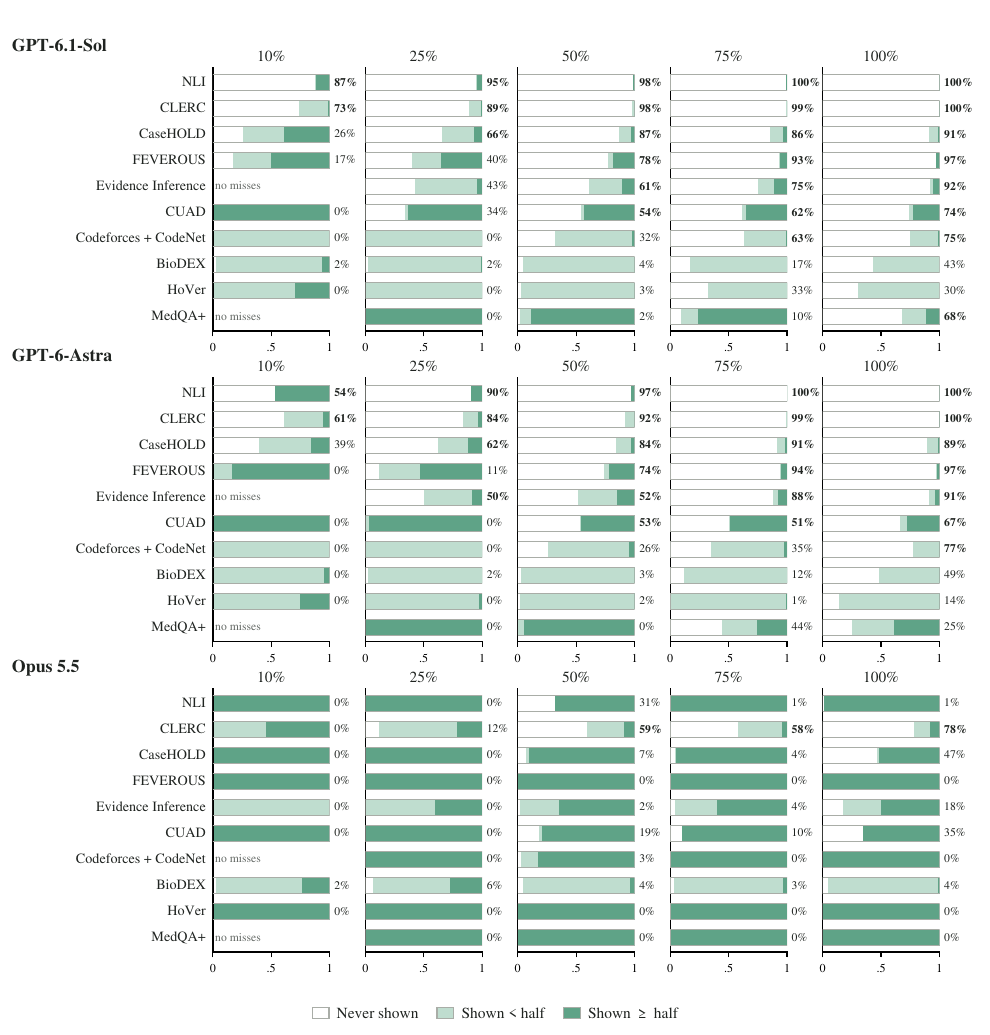}
  \caption{Missed answers by document exposure at every corpus size. As in Figure~\ref{fig:coverage}, for each agent at 10, 25, 50, 75, and 100\% of the full corpus. Missed answers are pooled over the runs at each size: three runs at 10, 25, 75, and 100\%, and the ten 50\% corpora. The figures listed are the \textit{never shown} documents percentage.}
  \label{fig:coverage-sizes}
\end{figure*}

\paragraph{Exposure across corpus sizes.}
Figure~\ref{fig:coverage-sizes} breaks down the exposure results from Figure~\ref{fig:coverage} across corpus sizes for each agent, and Table~\ref{tab:coverage-sizes} reports the percentage of documents \textit{mostly shown}.
For the Codex agents, the share of missed answers that were \textit{never shown} rises with corpus size on most tasks, while the share of documents \textit{mostly shown} falls.

\begin{table*}[!t]
  \centering\footnotesize\setlength{\tabcolsep}{3.5pt}
  \input{content/tables/tab_coverage_sizes.tex}
  \caption{Documents \textit{mostly shown} at every corpus size (\% of the corpus, mean).}
  \label{tab:coverage-sizes}
\end{table*}

\clearpage

%% file: content/tables/tab_coverage_sizes.tex
\begin{tabular}{@{}lrrrrrrrrrrrrrrr@{}}
\toprule
 & \multicolumn{5}{c}{GPT-6.1-Sol} & \multicolumn{5}{c}{GPT-6-Astra} & \multicolumn{5}{c}{Opus~5.5} \\
\cmidrule(lr){2-6} \cmidrule(lr){7-11} \cmidrule(lr){12-16}
Task & 10 & 25 & 50 & 75 & 100 & 10 & 25 & 50 & 75 & 100 & 10 & 25 & 50 & 75 & 100 \\
\midrule
NLI & 21 & 6.2 & 2.5 & 0.6 & 0.5 & 58 & 13 & 4.3 & 0.6 & 0.5 & 100 & 100 & 91 & 99 & 99 \\
CLERC & 1.3 & 0.9 & 0.2 & 0.1 & 0.1 & 3.0 & 2.5 & 0.7 & 0.1 & 0.1 & 46 & 14 & 5.7 & 4.2 & 5.8 \\
CaseHOLD & 25 & 4.7 & 2.8 & 2.6 & 1.1 & 15 & 9.1 & 2.9 & 1.6 & 1.5 & 100 & 100 & 81 & 86 & 26 \\
FEVEROUS & 33 & 23 & 11 & 5.1 & 3.1 & 62 & 44 & 13 & 4.0 & 2.4 & 99 & 98 & 97 & 95 & 92 \\
Evidence Inference & 6.5 & 12 & 11 & 9.1 & 4.7 & 8.3 & 13 & 15 & 7.2 & 4.6 & 24 & 45 & 44 & 53 & 42 \\
CUAD & 99 & 68 & 53 & 41 & 28 & 99 & 91 & 54 & 49 & 36 & 99 & 98 & 79 & 83 & 55 \\
Codeforces + CodeNet & 13 & 5.8 & 6.0 & 3.2 & 3.1 & 10 & 6.7 & 8.0 & 6.2 & 3.1 & 100 & 100 & 96 & 100 & 100 \\
MedQA+ & 100 & 91 & 87 & 74 & 41 & 100 & 100 & 82 & 42 & 41 & 100 & 100 & 100 & 100 & 100 \\
BioDEX & 2.3 & 0.5 & 0.1 & 0.1 & 0 & 2.1 & 0.2 & 0.1 & 0.1 & 0 & 12 & 15 & 1.9 & 1.3 & 0.4 \\
HoVer & 11 & 2.4 & 1.5 & 1.0 & 1.0 & 13 & 3.6 & 1.7 & 1.4 & 2.3 & 100 & 100 & 100 & 100 & 100 \\
\bottomrule
\end{tabular}

%% file: content/appendix/workflows.tex
\section{Stopping Audit}
\label{app:workflow}\label{app:stopping}

We inspect the final message from each of the 300 runs on 50\% corpora to check whether the agent acknowledges that its search may be incomplete, such as by noting unchecked documents or potentially missed answers.
Across the 200 Codex runs, 184 make no such acknowledgment (Table~\ref{tab:stopping}).
The remaining 16 runs (6 by GPT-6.1-Sol and 10 by GPT-6-Astra, covering 14 runs on NLI and 2 on FEVEROUS) explicitly note incomplete coverage, stating for example: ``Best-effort result: only 6,150 of 145,478 records were individually reviewed; exhaustive accuracy is unverified.''
By contrast, Opus~5.5 acknowledges potential incompleteness in 72 of its 100 runs and notes a methodological caveat in another 27.

\begin{table}[t]
  \centering\footnotesize\setlength{\tabcolsep}{4pt}
  \begin{tabular}{@{}lcccc@{}}
    \toprule
    Agent & Runs & Admits & Caveat only & Neither \\
    \midrule
    GPT-6.1-Sol & 100 & 6 & 0 & 94 \\
    GPT-6-Astra & 100 & 10 & 0 & 90 \\
    Opus~5.5 & 100 & 72 & 27 & 1 \\
    \bottomrule
  \end{tabular}
  \caption{Agent's final message for all 100 \dataset{} examples. \emph{Admits}: denotes that the agents states its result may be incomplete, or that some documents were left unchecked. \emph{Caveat only}: denotes that the agent only mentions a methodological limitation, such as keyword filtering or sampling. \emph{Neither}: the agent's final message does not mention any limitation.}
  \label{tab:stopping}
\end{table}

%% file: content/appendix/trajectories.tex
% ACL two-column appendix: each case study remains a full-width figure float.
% No \onecolumn / \captionof / subsection switches. Figure labels are preserved.
% Requires xcolor and caption (both are used by the original appendix).
\definecolor{trbg}{HTML}{F5F6F3}
\definecolor{trborder}{HTML}{D8DDD9}
% Keep single figures at the top of double-column float pages.
\makeatletter
\setlength{\@dblfptop}{0pt}
\setlength{\@dblfpsep}{16pt}
\setlength{\@dblfpbot}{0pt plus 1fil}
\makeatother
\newsavebox{\trsave}
\newenvironment{trsnippet}{%
  \par\smallskip\noindent\begin{lrbox}{\trsave}%
  \begin{minipage}{\dimexpr\linewidth-2\fboxsep-2\fboxrule\relax}\scriptsize%
}{%
  \end{minipage}\end{lrbox}%
  \fcolorbox{trborder}{trbg}{\usebox{\trsave}}\par\smallskip%
}
\newcommand{\trstep}[1]{\par\medskip\noindent\textbf{#1}\enspace}
\newcommand{\trhead}[4]{%
  \noindent{\normalsize\bfseries #1}\par\vspace{3pt}%
  \noindent{\footnotesize #2\quad\textbar\quad #3\hfill\textit{P/R/F1: #4}}\par\vspace{5pt}%
}

\section{Example Agent Trajectories}
\label{app:trajectories}

We examine seven agent trajectories illustrating the two strategies in Section~\ref{sec:analysis}: searching with rules (Section~\ref{sec:strategies}) and reading with sub-agents (Section~\ref{sec:claudecode}).
Figures~\ref{fig:traj-nli-rules}, \ref{fig:traj-code-rules}, and~\ref{fig:traj-feverous} show searching with rules; Figures~\ref{fig:traj-nli-opus}, \ref{fig:traj-code-opus}, \ref{fig:traj-medqa}, and~\ref{fig:traj-biodex} show reading with sub-agents, the last after a keyword screen.
The first two, Figures~\ref{fig:traj-nli-rules} and~\ref{fig:traj-nli-opus}, are the two runs of Figure~\ref{fig:pipeline}.
For each run, we show how the agent searches the corpus, what document text reaches the language model, and where answers are missed.
Code and trajectory excerpts are verbatim; \texttt{[...]} marks omissions.
Cost accounting appears in Appendix~\ref{app:cost}.

\begin{figure*}[p]
  \footnotesize
  \trhead{NLI: Searching with rules}{GPT-6.1-Sol (Codex)}{100\% corpus: 290,956 documents}{.63 / .18 / .28}
\noindent\textbf{Task.} Find every statement contradicted by its accompanying passage.

\trstep{Step 1. Write rules to screen the corpus.}
After inspecting a sample of records, Sol writes rules for negation, actions, dates, quantities, and other cues of contradiction. The program applies these rules across the corpus, while the language model sees only selected document text in tool outputs.
\begin{trsnippet}
\begin{verbatim}
Many statements add details that the passage leaves unknown, so I’m treating those as insufficient
information. I’m checking for direct conflicts such as incompatible actions, negation, dates,
quantities, and descriptions.
[...]
NEG=R(r"\b(no|not|never|nobody|nothing|none|neither|without|cannot|can't|doesn't|don't|didn't|isn't| [...]
ACTS={
'walk':r'\bwalk(?:s|ing|ed)?\b|\bstroll(?:s|ing|ed)?\b',
'sit':r'\bsitt?ing\b|\bsits?\b|\bsat\b|\bseated\b',
[...]
def classify(r):
[...]
 # Matching positive content within the scope of a negation.
 for rev,a,b in [(False,h,p),(True,p,h)]:
  for marker,ww,scope in neg_scopes(a):
   bset=cs(b);hits=scope&bset
[...]
   if len(hits)/len(scope)>=.65 and first in bset:
[...]
     add('neg_'+('p' if rev else 'h'),.93 if len(hits)>1 else .83)
[...]
  if 'sleep' in ha-pa and len(pa-{'sit','lie'}) and not r['negp']:add('sleep_active',.97)
[...]
 for r in rows:
  r=classify(r);out.append(r)
[...]
RULE COUNTS Counter({'neg_h': 1084, 'neg_p': 741, 'gender': 355, 'sleep_active': 182, [...]
candidates 3901
\end{verbatim}
\end{trsnippet}

\trstep{Step 2. Review sample matches.}
Sol prints samples of the matches and refines rules that mistake unsupported details for contradictions or confuse who acts and when. The model reviews these samples rather than reading every document.
\begin{trsnippet}
\begin{verbatim}
180174 [['neg_h', 0.83]]
P: Those who do venture here are rewarded with beautiful scenery and friendly people.
S: There are many places on Earth that do not have friendly people.
[...]
The review found false matches where a statement adds an unknown detail, changes who acts, or refers
to a different time. I’ve tightened those checks and am validating numerical conflicts against the
specific quantity named.
[...]
sample(range(260),[11,25,45,51,61,85,106,202,207,245])
# Independently reviewed rule samples.
base=[r for r in map(json.loads,open('/tmp/scored.jsonl')) if r['reasons']]
random.Random(80).shuffle(base)
sample([r['i'] for r in base[:120]],[8781,258669,281049,166047,169776,44448,137192,243629,151502, [...]
[...]
records 290956 answer_ids 3699 manual_reviews 745 positive_reasons Counter({'numeric_bound': 1979,
'neg_h': 398, 'gender': 340, 'review': 221, 'numeric_exact': 219, 'sleep_active': 117, [...]
\end{verbatim}
\end{trsnippet}

\trstep{Step 3. Write the answer.}
Sol returns the records selected by its rules and acknowledges that the classifications remain heuristic.
\begin{trsnippet}
\begin{verbatim}
 qualified=bool(rs)
 conf=.88 if qualified else .91
[...]
 if i in manual:
  qualified=manual[i];conf=.99;rs=[('review',.99)] if qualified else []
 if qualified:answers.append(r['id'])
[...]
answer_ids 3260 evidence_records 290956 reviewed 1285
[...]
"Classifications are heuristic and remain partly unverified."
\end{verbatim}
\end{trsnippet}

\noindent\textbf{Outcome.} Text from only 0.57\% of documents reaches the language model, and 99\% of missed answers are \emph{never shown}. The program processes every record, but its rules miss contradictions that the model never examines.
  \caption{\textbf{Searching with rules on NLI.} Sol applies rules across the corpus and reviews sample matches, leaving most missed answers unseen by the model.}
  \label{fig:traj-nli-rules}
\end{figure*}

\begin{figure*}[p]
  \footnotesize
  \trhead{NLI: Reading every record with sub-agents}{Opus 5.5 (Claude Code), Haiku 5.5 sub-agents}{100\% corpus: 290,956 documents}{.55 / .80 / .65}
\noindent\textbf{Task.} Find every statement contradicted by its accompanying passage.

\trstep{Step 1. Split the corpus into chunks.}
Opus parses all 290,956 records and writes them into 179 chunk files of about 340,000 characters, with 771 to 1,696 records each.
\begin{trsnippet}
\begin{verbatim}
cd /tmp/work_wad && python3 -I mkchunks.py 340000 [...]
[...]
179 [1627, 1619, 1578, 1608, 1627] 771 1696
\end{verbatim}
\end{trsnippet}

\trstep{Step 2. Write one set of instructions for every sub-agent.}
The instructions define three labels, forbid scripts, and ask for a question mark when the sub-agent is unsure. After a pilot on the first chunk, each of the other 178 Haiku sub-agents receives only its chunk ID, its number of records, and the path to the instructions.
\begin{trsnippet}
\begin{verbatim}
For EVERY record decide one label:
  C = the statement is DEFINITELY INCORRECT given the passage (the passage, plus ordinary world knowledge,
 refutes it)
  E = the passage supports the statement
  N = neither: the passage doesn't give enough information; the statement might be true
[...]
2. Judge every record yourself by actually reading it. Do NOT write scripts, keyword rules or any automated
 classifier — the labels must be your own judgment.
[...] Append "?" to the label (C?, N?, E?) when you are genuinely unsure.
[...]
Your chunk ID is c001 and it has N=1619 records. Read /tmp/work_wad/instructions.md and follow it exactly
 (substitute c001 for {C} and 1619 for {N}).
\end{verbatim}
\end{trsnippet}

\trstep{Step 3. Label about 300 records per read.}
The pilot sub-agent reads its chunk 900 lines, about 300 records, at a time and appends the labels for each read in one command. It labels record 855 as neither, although the dataset counts it as contradicted, and record 925 as an unsure contradiction, which the dataset does not count. After checking a sample of the pilot's labels, Opus runs the other 178 chunks in waves.
\begin{trsnippet}
\begin{verbatim}
[855] S: The photographer's arms are full of groceries.
P: A photographer is photographing a vine covered building.
[...]
855 N
[...]
[925] S: The young boy is not having fun.
P: A young boy in a red shirt plays in a tree.
[...]
925 C?
[...]
     60 C
     35 C?
    908 E
    163 E?
    459 N
      2 N?
\end{verbatim}
\end{trsnippet}

\trstep{Step 4. Merge the labels and keep the unsure contradictions.}
Opus merges the 179 label files and checks that every record has exactly one label. It prints 30 random unsure contradictions, judges about 23 of them real, and returns all 16,888 records labeled C or C?.
\begin{trsnippet}
\begin{verbatim}
Counter({'E': 156493, 'N': 59868, 'E?': 35765, 'N?': 21942, 'C': 11727, 'C?': 5161}) 290956
[...]
All 290,956 records are labeled with no gaps: 11,727 C plus 5,161 low-confidence C?. Before deciding
 whether to include the C? items, I'll spot-check a sample of them.
[...]
- 5,161 unsure contradictions. In a sample of 30 I checked, about 23 were real, so including them should
 help more than hurt.
\end{verbatim}
\end{trsnippet}

\noindent\textbf{Outcome.} Every record reaches a sub-agent, and 288,540 of the 290,956 are \textit{mostly shown}. By the dataset labels, 67.5\% of the C labels are answers but only 26.4\% of the C? labels, so keeping the C? labels raises recall from 68.0 to 79.7 and lowers precision from 67.5 to 54.9. The sub-agents labeled 1,743 of the 2,359 missed answers neither and 616 supported, and 2,345 of the missed answers were \textit{mostly shown}.
  \caption{\textbf{Reading with sub-agents on NLI.} Opus has Haiku sub-agents label every record, then keeps the unsure contradictions after checking 30 of them, which raises recall but lowers precision.}
  \label{fig:traj-nli-opus}
\end{figure*}

\begin{figure*}[p]
  \footnotesize
  \trhead{Codeforces + CodeNet: Screening with sample tests}{GPT-6.1-Sol (Codex)}{100\% corpus: 6,109 submissions}{.68 / .22 / .33}
\noindent\textbf{Task.} Find every submission the online judge rejected for a wrong answer or a runtime error. The hidden tests are not provided.

\trstep{Step 1. Run the sample tests.}
Sol runs submissions against the sample tests in their problem statements. Programs that pass the available examples receive a passing status from the screening program.
After counting the documents, Sol states its plan:
\begin{trsnippet}
\begin{verbatim}
There are 6,109 records. I’ll run the published examples first, then review failures and check for bugs
those examples may miss. [...]
\end{verbatim}
\end{trsnippet}
\begin{trsnippet}
\begin{verbatim}
def run(r):
    result = {'id': r['id'], 'file':r['file'], 'title':r['title'], 'tests':[]}
    ex=examples(r['statement'])
    if not ex:
        result['status']='no_samples';return result
    [...]
    for i,(inp,expected) in enumerate(ex):
        t={'index':i,'input':inp,'expected':expected}
        [...]
            if p.returncode:t.update(status='error',err=p.stderr.decode(errors='replace')[-2000:])
            elif same(out,expected):t['status']='pass'
            else:t['status']='mismatch'
    [...]
    result['status']='pass' if all(t['status']=='pass' for t in result['tests']) else 'review'
\end{verbatim}
\end{trsnippet}
The screen's last progress lines report the three groups:
\begin{trsnippet}
\begin{verbatim}
6000 {'pass': 4023, 'no_samples': 1157, 'review': 820} 63.7
DONE {'pass': 4091, 'no_samples': 1183, 'review': 835}
\end{verbatim}
\end{trsnippet}

\trstep{Step 2. Review failures and suspects.}
Sol reviews submissions flagged by the sample tests and searches for patterns such as deep recursion and undefined names. It records verdicts for the submissions it investigates.
\begin{trsnippet}
\begin{verbatim}
python /tmp/decide.py 'Recursion reaches beyond the default limit on a legal long path.' d2ebf8f0f [...] 
python /tmp/decide.py 'Uses unquoted No, causing NameError on e.g. N=44.' dfed7fabd
python /tmp/decide.py 'Calls undefined encode_gray and decode_gray functions.' daa30f425
python /tmp/decide.py 'Uses undefined n when input is already a palindrome.' d286f9e74
\end{verbatim}
\end{trsnippet}
The helper it wrote for these verdicts stores only positive decisions:
\begin{trsnippet}
\begin{verbatim}
if len(sys.argv)>2:
 reason=sys.argv[1]
 for term in sys.argv[2:]:
  r=next(r for r in rs if term in r['file'] or term==r['id'])
  ds[r['id']]={'qualifies':True,'confidence':0.99,'file':r['file'],'title':r['title'],'reason':reason}
 path.write_text(json.dumps(ds,indent=2))
print('qualifying',sum(d['qualifies'] for d in ds.values()))
\end{verbatim}
\end{trsnippet}
One of its review queues collects long programs that have no sample tests:
\begin{trsnippet}
\begin{verbatim}
 a=[]
 for r in rs:
  if st[r['id']]=='no_samples' and r['lang']!='C++' and len(r['code'].splitlines())>25:
   a.append(r|{'compact':compact(r['code'])})
 json.dump(a,open('/tmp/no_samples_long.json','w'))
\end{verbatim}
\end{trsnippet}
Midway through the review, it notes:
\begin{trsnippet}
\begin{verbatim}
I’m continuing the edge-case review, focusing on input handling, recursion depth, and constructed
answers. Some submissions pass the examples but fail when an ordinary input value changes.
\end{verbatim}
\end{trsnippet}

\trstep{Step 3. Write the answer.}
Sol returns the submissions it identified as rejected. Submissions that passed the sample tests generally do not reach its review.
Its final script gives every unflagged record a confidence based on its sample-test status:
\begin{trsnippet}
\begin{verbatim}
answer=[r['id'] for r in records if decisions.get(r['id'],{}).get('qualifies',False)]
evidence={}
for r in records:
 positive=r['id'] in answer
 status=results[r['id']]['status']
 confidence=decisions[r['id']]['confidence'] if positive else (0.94 if status=='pass' else 0.84 [...]
 evidence[r['id']]={'qualifies':positive,'confidence':confidence,'files':[r['file']]}
\end{verbatim}
\end{trsnippet}
\begin{trsnippet}
\begin{verbatim}
"Created answer.txt with 97 record IDs [...]"
\end{verbatim}
\end{trsnippet}

\noindent\textbf{Outcome.} Of the 221 rejected submissions that pass every sample test, Sol misses 197. Passing the samples does not establish correctness on hidden tests, so the program leaves many rejected submissions outside the model's review.
  \caption{\textbf{Screening with sample tests on Codeforces + CodeNet.} Sol reviews failures on the provided examples but misses submissions rejected on hidden tests.}
  \label{fig:traj-code-rules}
\end{figure*}

\begin{figure*}[p]
  \footnotesize
  \trhead{Codeforces + CodeNet: Reading every submission}{Opus 5.5 (Claude Code), Haiku 5.5 sub-agents}{100\% corpus: 6,109 submissions}{.59 / .96 / .73}
\noindent\textbf{Task.} Find every submission the online judge rejected for a wrong answer or a runtime error. The hidden tests are not provided.

\trstep{Step 1. Run the sample tests and batch every submission.}
Like Sol, Opus first runs every submission on the sample tests in its statement: 3,939 pass all of them and 1,183 have none. After reclassifying 124 whitespace and formatting mismatches, it assigns all 6,109 submissions to 104 batches: 29 of about 30 for the 863 whose output still differs from a sample or that crash or hang, and 75 of about 70 for the other 5,246.
\begin{trsnippet}
\begin{verbatim}
Counter({'allok': 3939, 'noex': 1183, ('WA',): 601, ('OK', 'WA'): 341, ('RE',): 30, ('TO',): 8,
 ('OK', 'RE'): 3, ('OK', 'TO'): 3, ('TO', 'WA'): 1})
[...]
for c in chunk(mis,30): batches.append(('M',c))
for c in chunk(pas,70): batches.append(('P',c))
[...]
863 5246
104 29
\end{verbatim}
\end{trsnippet}

\trstep{Step 2. Ask the sub-agents to read the code against the statement.}
Each batch goes to one Haiku sub-agent. Its prompt names only the batch file, which lists each submission with its sample-test status, and a shared instruction file. The instructions say that passing the samples is not evidence of correctness, list bugs to look for, and ask for brute-force comparisons on random inputs when reading leaves doubt; slowness alone is not a reason to flag, as no submission was rejected for time.
\begin{trsnippet}
\begin{verbatim}
[...] IMPORTANT: most failing submissions in this corpus PASS the examples - the bug shows only on hidden
 tests. So "sample OK" is not evidence of correctness; read the code against the statement.
[...]
For each record: read the statement and the code [...]. Look for: wrong algorithm / wrong greedy, missed
 edge cases (n=1, all equal, zero, negatives, empty answer, maximum values), off-by-one, [...] integer
 overflow in C++, recursion depth, [...]. When suspicious but unsure, actually test: write a brute-force
 solver for small inputs and compare on random tests [...]
\end{verbatim}
\end{trsnippet}

\trstep{Step 3. Read each submission and test the doubtful ones.}
After two pilot batches, Opus adds that a failing input confirmed by a brute force or by hand means a rejection, and runs the other 102 batches in waves of 20, the limit on concurrent sub-agents. In batch b090, a sub-agent reads a Taxi submission that passes both samples but divides the total number of children by four, although each group must ride in one taxi. It runs the program on groups of 3, 3, and 2 children and records the failure. The judge rejected this submission, which passes Sol's screen.
\begin{trsnippet}
\begin{verbatim}
=== dd02302ec.txt | r586e09f60 | Python 3, CPython | Taxi
    STATUS: sample1: OK; sample2: OK
[...]
[...] What minimum number of cars will the children need if all members of each group should ride in the
 same taxi (but one taxi can take more than one group)?
[...]
a = sum(s)

if a%4 == 0:
    print(a//4)
else:
    print(a//4+1)
[...]
[...] "r586e09f60|true|0.97|Outputs ceil(total/4); wrong for groups 3,3,2 (needs 3 taxis, prints 2)" [...]
\end{verbatim}
\end{trsnippet}

\trstep{Step 4. Merge the verdicts and recheck environment flags.}
Opus merges the 104 verdict files and checks that every submission has exactly one verdict; the sub-agents flag 518. It then rereads the flags that depend on how the judge runs code, such as letter case, recursion depth, blank lines, and compile tricks. It clears 20 flags, one of them on a rejected submission, adds 3 bugs that sub-agents found but did not flag, and writes 501 submissions.
\begin{trsnippet}
\begin{verbatim}
missing 0 [] extra []
flagged 518
[...]
set_false = {
[...]
 'r887618dfe':'AtCoder numba AOT ONLINE_JUDGE compile trick works on judge',
[...]
for k in atcoder_stack: set_false[k]='AtCoder large stack: deep recursion with raised limit OK'
[...]
501
\end{verbatim}
\end{trsnippet}
\noindent Its final message expects some failures to remain:
\begin{trsnippet}
\begin{verbatim}
- **Missed failures:** some subtle bugs in records that passed their examples were probably not caught,
 so the true number of failures is likely somewhat higher than 501.
\end{verbatim}
\end{trsnippet}

\noindent\textbf{Outcome.} Every submission is assigned to a sub-agent, and 6,078 of the 6,109 are \textit{mostly shown}. Of the 221 rejected submissions that pass Sol's sample-test screen, Opus returns 214, while Sol returns 24. All 11 missed answers were \textit{mostly shown} but not returned: the other 7 sample-passing rejections and 4 submissions without sample tests. Reading every submission raises recall to .96, but 207 of the 501 returned submissions were accepted by the judge.
  \caption{\textbf{Reading with sub-agents on Codeforces + CodeNet.} Opus runs the sample tests as Sol does, but its sub-agents read every submission against its statement and find most of the rejections that the samples miss.}
  \label{fig:traj-code-opus}
\end{figure*}

\begin{figure*}[p]
  \footnotesize
  \trhead{FEVEROUS: Word-overlap screening}{GPT-6-Astra (Codex)}{50\% corpus: 25,252 claims}{.48 / .60 / .53}
\noindent\textbf{Task.} Find every claim refuted by its accompanying Wikipedia evidence.

\trstep{Step 1. Index unmatched words.}
Astra writes a program that records content words in each claim that do not appear in its evidence. It uses word overlap to screen claims before reviewing selected candidates.
\begin{trsnippet}
\begin{verbatim}
stop=set('a an the of at in on by to and or for as with from that which who whom whose was were is [...]
[...]
def toks(t): return re.findall(r'[a-z]+|\d+(?:\.\d+)?',norm(t))
def content(t):return [cv(w) for w in toks(t) if w not in stop]
[...]
ct=content(c); et=set(content(e)); miss=list(dict.fromkeys(x for x in ct if x not in et))
records.append({'file':p.name,'id':label, [...] 'missing':miss,'ratio':len(miss)/max(1,len(set(ct)))})
[...]
claims 25252 exact 7022
missing counts [(0, 7022), (1, 5962), (2, 4134), (3, 2758), (4, 1794), (5, 1181), (6, 736), [...]
\end{verbatim}
\end{trsnippet}

\trstep{Step 2. Review candidate lists.}
Astra filters the index into candidate lists based on numbers, negation, countries, and other cues. It prints candidates in slices and records verdicts for the claims it reviews.
\begin{trsnippet}
\begin{verbatim}
ns=[w for w in r['missing'] if w[0].isdigit() and float(w)>31]
if ns and r['id'].split()[1] not in man and r['ratio']<.35:
[...]
NUMERIC CANDIDATES 817
[...]
# Suppress connective and common words that routinely express table information.
generic='win play while but receiv more finish where onli maximum such after made game against [...]
[...]
hard=[w for w in r['missing'] if w not in generic and not w[0].isdigit() and len(w)>2]
if not hard:continue
if r['id'].split()[1] in man:continue
[...]
HARD CANDIDATES 6703
\end{verbatim}
\end{trsnippet}
Verdicts are recorded with \texttt{label.py}, here after one slice of the numeric list:
\begin{trsnippet}
\begin{verbatim}
label=sys.argv[1]=='1'
for x in sys.argv[2:]:d[x]={'qualifies':label,'confidence':0.99}
[...]
python /tmp/claim_review/subshow.py numeric 80 125
[...]
92 #40093 Cees Lok(born 18 August 1966) became the coach after Johan Neeskens was fired and [...]
E: [Cees Lok] / Section: Managerial career > NEC / After Johan Neeskens was fired as head coach at [...]
93 #94370 Alvdal IL (full name Alvdal Idrettslag) was founded in 1879.
E: [Alvdal IL] / Alvdal Idrettslag is a Norwegian sports club from Alvdal, founded in 1902.
[...]
python /tmp/claim_review/label.py 1 74742 94370 2938 12160 5439 29498 38689 52647 44381 [...]
[...]
Reviewed 806 refuted 561
[...]
Reviewed 2467 refuted 1253
\end{verbatim}
\end{trsnippet}

\trstep{Step 3. Write the answer.}
Astra labels claims without a recorded verdict as not refuted and assigns confidence using word overlap, including for claims never shown to the model.
Early in the run, Astra notes that every claim needs an entry; its final script fills in the claims without a verdict:
\begin{trsnippet}
\begin{verbatim}
The corpus contains 25,252 claims. The question also requires an `evidence.json` entry for every [...]
\end{verbatim}
\end{trsnippet}
\begin{trsnippet}
\begin{verbatim}
# These collections were checked against their complete selected evidence.
checked=set()
[...]
decision=m.get(n)
if decision is not None:
  qualifies=decision['qualifies']
  confidence=decision['confidence']
  if qualifies and int(n) in contextual:confidence=0.84
elif ident in checked:
  qualifies=False;confidence=0.95
else:
  qualifies=False
  # Exact lexical coverage is strong support, while partial coverage leaves
  # more uncertainty about whether an unverified detail is a contradiction.
  confidence=0.9 if not r['missing'] else 0.78 if r['ratio']<0.25 else 0.72
[...]
Validated 1252 refuted claim labels and 25252 evidence entries.
\end{verbatim}
\end{trsnippet}

\noindent\textbf{Outcome.} Word overlap can miss contradictions: one claim swaps two reviewers' ratings, although both numbers appear in its evidence. Of the 404 missed answers, 269 are \emph{never shown} to the model.
  \caption{\textbf{Word-overlap screening on FEVEROUS.} Astra filters claims using unmatched words, then assigns default labels to claims it has not reviewed.}
  \label{fig:traj-feverous}
\end{figure*}

\begin{figure*}[p]
  \footnotesize
  \trhead{MedQA+: Sub-agents agree on false positives}{Opus 5.5 (Claude Code), Haiku 5.5 sub-agents}{50\% corpus: 2,338 notes}{.51 / .97 / .66}
\noindent\textbf{Task.} Find every clinical note containing an error in diagnosis, cause, treatment, or management.

\trstep{Step 1. Assign every note.}
Opus splits the notes into 20 chunks and sends each chunk to a Haiku sub-agent. All 20 prompts say that about half of the notes contain an error; the answer key marks 90 of the 2,338 notes (3.8\%).
\begin{trsnippet}
\begin{verbatim}
[...] split -n l/20 -d -a 2 all_files.txt list_ && for l in list_*; do [...]
[...]
You are a physician-level medical reviewer. Read the file [...]/chunk_00.txt (read ALL of it, using the
 Read tool in parts if needed; it is ~65KB). [...] Each clinical text is either fully correct, or
 contains exactly ONE medical error (in the style of the MEDEC benchmark: [...]). Roughly a large
 fraction (maybe ~half) of texts contain an error; judge each one independently on its merits.
\end{verbatim}
\end{trsnippet}

\trstep{Step 2. Collect first-pass flags.}
The sub-agents return a verdict for every note and flag 241. Opus prints samples of the verdicts and the reasons for all 241 flags.
\noindent A first-pass sub-agent flags a psychiatric note; Opus checks that every note has a verdict and samples notes marked correct with low confidence.
\begin{trsnippet}
\begin{verbatim}
{"file": "d1e005f89.txt", "id": "mc12b64842", "qualifies": true, "confidence": 0.7, "reason": "Only
 persecutory and Capgras-type delusions for 10 years, well groomed, no hallucinations or functional
 decline = delusional disorder; schizophrenia requires two Criterion A symptoms and decline, so the
 stated dx is wrong."}
[...]
print('missing', len(allf-set(r['file'] for r in rs)), 'extra', len(set(r['file'] for r in rs)-allf))
[...]
2338 2338
missing 0 extra 0
[...]
low=[r for r in rs if not r['qualifies'] and r['confidence']<=0.5]
\end{verbatim}
\end{trsnippet}

\trstep{Step 3. Run a second full pass.}
Opus reshuffles all 2,338 notes into 20 new chunks for 20 new sub-agents. The new prompt states that the error rate is unknown. The second pass flags 195 notes; 174 are flagged in both passes.
\begin{trsnippet}
\begin{verbatim}
random.seed(2026); random.shuffle(fs)
[...]
Each text is either entirely correct, or exactly ONE sentence has been altered to state a
 wrong-but-plausible answer (a distractor) [...]. The fraction of altered texts is unknown.
\end{verbatim}
\end{trsnippet}
\noindent Opus compares the passes and announces which notes it will read.
\begin{trsnippet}
\begin{verbatim}
both 174 only1 67 only2 21 p2 total 195
The second pass flagged 195. The two passes agree on 174 records, and they disagree on 88: 67 were
 flagged only by the first pass and 21 only by the second. I'll read those 88 myself.
\end{verbatim}
\end{trsnippet}

\trstep{Step 4. Merge the passes.}
Opus decides the 88 disagreements itself and keeps 10. It reads the 73 agreed flags on which either pass was unsure, and 10 of the other 101, and clears 19. In a third pass, four sub-agents review 393 notes that both passes called correct, with at least one pass assigning low confidence ($\le 0.5$); they flag 23, and Opus keeps 7. All other agreed flags stay in the answer.
\noindent Opus reads the psychiatric note from Step 2, which the answer key labels correct, and does not clear it.
\begin{trsnippet}
\begin{verbatim}
## d1e005f89.txt mc12b64842 0.7 0.5 P2: Ten years of isolated persecutory and Capgras-type delusions
 with intact functioning fits delusional disorder; paranoid schizophrenia is a poorer fit.
A 60 year old man is brought to a psychiatrist with a 10 year history, that he suspects his neighbours
 and he feels that whenever he passes by they sneeze and plan agaist behind his back. [...] Based on
 the findings, the patient was diagnosed with paranoid schizophrenia.
[...]
unflag=set('m26cc49215 m7ba61d339 mde7e6d7f5 m0b4cd14ac m550e555ba m4dddb7b41 m5c6eb6070 mc0647724b [...]
[...]
  if f in both and i not in unflag:
    q=True; conf=round((a[f]['confidence']+b[f]['confidence'])/2,2)
  elif i in add_dis: q=True; conf=add_dis[i]
  elif i in add_p3: q=True; conf=add_p3[i]
\end{verbatim}
\end{trsnippet}
\noindent Opus reports its answer with two caveats: some flags are uncertain, and a few errors may be missed.
\begin{trsnippet}
\begin{verbatim}
I found 172 of the 2,338 records with an error. [...]
[...]
- **Hard to settle from the text alone:** about 30–40 of the 172 depend on what the original exam's
 answer key intended, so some calls there could go either way. [...]
- **Possible misses:** the overlap between the two passes suggests a few more errors could still sit
 among the records called correct, but the third pass found only a few.
\end{verbatim}
\end{trsnippet}

\noindent\textbf{Outcome.} Every note is shown to two sub-agents, but 71 of the 85 false positives are flagged in both passes; the other 14 come from the 17 notes Opus adds from the disagreements and the third pass. Reading the full corpus supports high recall; merging the sub-agents' judgments still produces low precision.
  \caption{\textbf{Reading with sub-agents on MedQA+.} Two passes cover every note but retain false positives on which the sub-agents agree.}
  \label{fig:traj-medqa}
\end{figure*}

\begin{figure*}[p]
  \footnotesize
  \trhead{BioDEX: Keyword screening and excerpts}{Opus 5.5 (Claude Code), Haiku 5.5 sub-agents}{50\% corpus: 3,554 articles}{.47 / .71 / .55}
\noindent\textbf{Task.} For each requested adverse reaction, find every article in which the reaction was part of an adverse drug event.

\trstep{Step 1. Screen with keyword rules.}
Opus searches the corpus for terms naming the target reactions and retains matching articles for sub-agent review.
\begin{trsnippet}
\begin{verbatim}
"1": re.compile(r"acute kidney injur|acute renal (failure|injur)|acute kidney failure|\bAKI\b", re.I),
"2": re.compile(r"neutropeni", re.I),
"3": re.compile(r"thrombocytopeni", re.I),
[...]
    for q, p in pats.items():
        hits = [s for s in sents if p.search(s)]
[...]
        if v["ta"] or v["nstrong"] > 0 or (q == "4" and v["n"] > 0):
\end{verbatim}
\end{trsnippet}

\trstep{Step 2. Send matching excerpts to sub-agents.}
For each candidate, Opus sends its title, a truncated abstract, and the first seven matching sentences. The sub-agents are told to open the full article only when the excerpts are ambiguous.
\begin{trsnippet}
\begin{verbatim}
            qs.append((q, len(sel2), [trunc(s, strong[q]) for s in sel2[:7]]))
[...]
        txt += f"[Q{q} {names[q]}] candidate; body sentences matching: {n}" + (" (showing first 7)"
 if n>7 else "") + "\n"
[...]
Open the source file (/tmp/waa7d3d258b/pages/<file>) only when the snippets are genuinely ambiguous [...]
Reply only with "done: N records" plus any problems.
\end{verbatim}
\end{trsnippet}
Each sub-agent is first told to read a guidelines file that Opus wrote:
\begin{trsnippet}
\begin{verbatim}
- The reaction is the underlying disease / indication being treated, or pre-existed before the drug
  (e.g., ITP treated with romiplostim; patient admitted with fever from an infection and then given an
[...]
- Q3: Heparin-induced thrombocytopenia ~0.6; drug-induced immune thrombocytopenia ~0.75; [...]
[...]
Each record shows the title, abstract (truncated), and up to 7 body sentences matching each candidate
term. Usually that is enough. [...]
\end{verbatim}
\end{trsnippet}

\trstep{Step 3. Judge from the excerpts.}
In one answer article, the excerpt associates thrombocytopenia with COVID-19, but an omitted sentence also discusses a possible drug-induced reaction. The sub-agent does not open the article and assigns it a low score.
\begin{trsnippet}
\begin{verbatim}
=== RECORD r6ed7597b3 [...]
TITLE: Treatment with eltrombopag of severe immune thrombocytopenia and hemolytic anemia associated with
 COVID-19 pneumonia: a case report.
[...]
[Q3 Thrombocytopenia] candidate; body sentences matching: 18 (showing first 7)
  - Herein, we report a case of COVID-19 pneumonia associated with two severe blood diseases, immune [...]
  [...]
  - COVID-19 can be present together with immune thrombocytopenia and hemolytic anemia.
[...]
"r6ed7597b3": {"3": 0.15, "4": 0.1},
[...]
done: 82 records
\end{verbatim}
\end{trsnippet}
The omitted passage, quoted from the article (not from the trajectory):
\begin{trsnippet}
\begin{verbatim}
We must consider, however, that the patient was treated with azithromycin and we cannot exclude the
 possibility of drug-induced thrombocytopenia and anemia. However, azithromycin rarely causes
 immune-mediated anemia and thrombocytopenia. [...]
\end{verbatim}
\end{trsnippet}

\trstep{Step 4. Apply score cutoffs.}
Opus requires articles to hit a minimum score for each reaction. This sample only gets a 0.15, missing the 0.4 requirement. As a result, the article is discarded, and the full article is never opened by any model.
\begin{trsnippet}
\begin{verbatim}
    ps=sorted([v[q] for v in r.values() if q in v],reverse=True)
    G=sum(ps); best=None
    for t in [.25,.3,.35,.4,.45,.5,.55,.6]:
        S=[p for p in ps if p>=t]; f=2*sum(S)/(len(S)+G)
[...]
   r6ed7597b3 0.15 Treatment with eltrombopag of severe immune thrombocytopenia [...]
[...]
ans={q:sorted(k for k,v in r.items() if q in v and v[q]>=thr[q]) for q in '12345'}
[...]
3. **Cutoffs.** I set them where the expected F1 score was highest: 0.4 for questions 1–3 and 0.35 for
 questions 4–5. My spot checks of borderline cases and of low-scored articles with the term in the title
 looked sound.
\end{verbatim}
\end{trsnippet}

\noindent\textbf{Outcome.} Most missed answers pass the keyword screen but are scored below the cutoff. The sub-agents see matching excerpts rather than full articles, and may miss evidence that appears later in the text.
  \caption{\textbf{Keyword screening and excerpts on BioDEX.} Sub-agents judge candidates from partial text, so relevant evidence can remain unseen.}
  \label{fig:traj-biodex}
\end{figure*}